\documentclass{article}

     \PassOptionsToPackage{numbers, compress}{natbib}

\usepackage[preprint]{neurips_2018}
 \usepackage{hyperref}
\usepackage{url}

\usepackage{amsmath,amsfonts,amssymb,amsthm}

\usepackage{prettyref}
\usepackage{mathrsfs}
\usepackage{graphicx}
\usepackage{wrapfig}
\usepackage{subfig}
\usepackage{bbold}
\usepackage{tabu}
\usepackage{MnSymbol}
\usepackage{multirow}
\usepackage{booktabs}
\usepackage{enumitem}
\usepackage{algorithm}
\usepackage[table]{xcolor}
\usepackage[noend]{algpseudocode}

\newtheorem{theorem}{\TE{Theorem}}[section]
\newtheorem{property}{\TE{Property}}[section]
\newtheorem{define}[theorem]{\TE{Definition}}
\newtheorem{lemma}[theorem]{\TE{Lemma}}

\newtheorem{corollary}[theorem]{\TE{Corollary}}
\algnewcommand{\IfThenElse}[3]{
  \State \algorithmicif\ #1\ \algorithmicthen\ #2\ \algorithmicelse\ #3}
  
\algnewcommand{\LineComment}[1]{\State \(\triangleright\) #1}
\algdef{SE}[DOWHILE]{Do}{doWhile}{\algorithmicdo}[1]{\algorithmicwhile\ #1}
\newcommand*{\colorboxed}{}
\def\colorboxed#1#{%
  \colorboxedAux{#1}%
}
\newcommand*{\colorboxedAux}[3]{%
  \begingroup
    \colorlet{cb@saved}{.}%
    \color#1{#2}%
    \boxed{%
      \color{cb@saved}%
      #3%
    }%
  \endgroup
}

\newrefformat{fig}{Figure~\ref{#1}}
\newrefformat{par}{Section~\ref{#1}}
\newrefformat{appen}{Appendix~\ref{#1}}
\newrefformat{sec}{Section~\ref{#1}}
\newrefformat{sub}{Section~\ref{#1}}
\newrefformat{table}{Table~\ref{#1}}
\newrefformat{ass}{Assumption~\ref{#1}}
\newrefformat{alg}{Algorithm~\ref{#1}}
\newrefformat{def}{Definition~\ref{#1}}
\newrefformat{thm}{Theorem~\ref{#1}}
\newrefformat{cor}{Corollary~\ref{#1}}
\newrefformat{lem}{Lemma~\ref{#1}}
\newrefformat{step}{Step~\ref{#1}}
\newrefformat{prop}{Property~\ref{#1}}
\newrefformat{ln}{Line~\ref{#1}}
\newrefformat{rem}{Remark~\ref{#1}}
\newrefformat{eq}{Equation~\ref{#1}}
\newrefformat{pb}{Problem~\ref{#1}}
\newrefformat{it}{Item~\ref{#1}}
\newrefformat{te}{Term~\ref{#1}}
\def\Eqref Eq:#1:{\eqref{eq:#1}}
\newrefformat{Eq}{Equation~\Eqref#1:}

\newcommand{\TE}[1]{\textbf{#1}}

\newcommand{\ResizedEq}[1]{\resizebox{0.94\hsize}{!}{$\begin{aligned}#1\end{aligned}$}}

\newcommand{\FPP}[2]{\frac{\partial{#1}}{\partial{#2}}}
\newcommand{\FPPR}[2]{{\partial{#1}}/{\partial{#2}}}
\newcommand{\FPPT}[2]{\frac{\partial^2{#1}}{\partial{#2}^2}}

\newcommand{\FPPTT}[3]{\frac{\partial^2{#1}}{\partial{#2}\partial{#3}}}

\newcommand{\TWO}[2]{\left(\setlength{\arraycolsep}{1pt}\begin{array}{cc}{#1}, & {#2}\end{array}\right)}

\newcommand{\argmin}[1]{\underset{#1}{\text{argmin}}\;}

\newcommand{\CFree}{\mathcal{C}_{\text{free}}}
\newcommand{\CObs}{\mathcal{C}_{\text{obs}}}
\newcommand{\CAll}{\mathcal{C}}
\newcommand{\CH}{\text{CH}}

\newcommand{\PXInv}[1]{\frac{1}{(#1)^+}}
\newcommand{\PXInvR}[1]{1/(#1)^+}

\usepackage{pifont}
\newcommand{\cmark}{\textcolor{green}{\ding{51}}}%
\newcommand{\xmark}{\textcolor{red}{\ding{55}}}
\usepackage[export]{adjustbox}

\title{Efficient Differentiable Contact Model\\ 
with Long-range Influence}

\author{Xiaohan Ye$^1$, Kui Wu$^2$, Zherong Pan$^{2\dagger}$, and Taku Komura$^{1\dagger}$  \\
\thanks{$^\dagger$ indicates corresponding author. $^2$independent researcher. \{walker.kui.wu@gmail.com, zherong.pan.usa@gmail.com\}. $^1$The Department of Computer Science, Hong Kong University. \{u3010417@connect.hku.hk, taku@cs.hku.hk\}}}

\author{Xiaohan Ye, Kui Wu, Taku Komura, Zherong Pan}

\begin{document}

\maketitle

\begin{abstract}
With the maturation of differentiable physics, its role in various downstream applications—such as model-predictive control, robotic design optimization, and neural PDE solvers—has become increasingly important. However, the derivative information provided by differentiable simulators can exhibit abrupt changes or vanish altogether, impeding the convergence of gradient-based optimizers. In this work, we demonstrate that such erratic gradient behavior is closely tied to the design of contact models. We further introduce a set of properties that a contact model must satisfy to ensure well-behaved gradient information. Lastly, we present a practical contact model for differentiable rigid-body simulators that satisfies all of these properties while maintaining computational efficiency. Our experiments show that, even from simple initializations, our contact model can discover complex, contact-rich control signals, enabling the successful execution of a range of downstream locomotion and manipulation tasks.
\end{abstract}
\section{Introduction}
Recent advancements in differentiable physical models~\citep{Werling-RSS-21,huang2024differentiable} have unlocked a range of downstream applications, including model-based reinforcement learning~\cite{xu2022accelerated}, shooting-based controller optimization~\citep{amos2018differentiable}, and robot co-design~\citep{Xu-RSS-21}. State-of-the-art models now extend to various material types, encompassing both rigid and deformable bodies, while offering first-order gradient information. A notable advantage of differentiable physics is its ability to automatically discover contact-rich motions from trivial initializations~\citep{mordatch2012discovery,pang2023global}. Achieving this, however, requires an ideal contact model that strikes a balance between accurately approximating physical contact mechanisms and providing meaningful gradient information. Over the years, significant efforts have been made to optimize this balance~\citep{Werling-RSS-21,huang2024differentiable,10105986}.
Despite their significant progress, recent systematic analyses~\citep{antonova2023rethinking,suh2022differentiable,suh2022bundled} have highlighted several pitfalls in the gradient information provided by differentiable physics systems. While analytic gradients are beneficial across much of the objective landscape, they can exhibit rugged behavior when optimizing over non-smooth interactions and may vanish in nearly flat regions. Consequently, optimizers are often prone to becoming trapped in undesirable local optima.

To address these challenges, existing techniques~\citep{antonova2023rethinking,li2022contact} employ global optimization algorithms, such as Bayesian optimization and optimal transportation, to escape local optima. While we agree that global search methods are crucial for complementing local gradient-based optimization, we argue that many of the issues with poor gradient information can be mitigated by improving the contact models within existing differentiable physics frameworks. Specifically, both rugged and vanishing gradients stem from the contact model itself. When two rigid objects come into contact, the abrupt introduction of contact forces results in rugged gradients. Conversely, in the absence of contact, the lack of direct interaction leads to vanishing gradients. 

In this paper, we make both theoretical and practical contributions, both aimed at enhancing the gradient landscape of differentiable rigid body simulators. Theoretically, we introduce in~\prettyref{sec:property} a set of properties that define a well-behaved contact model. These properties ensure that the contact model supports differentiation and can provably prevent inter-penetration, generate physically plausible contact forces, and provide non-vanishing gradients even when objects are arbitrarily far apart. The last property leads to long-range influence, allowing us to discover novel contact-rich motions, even from a trivial initialization where objects are distant from each other. Practically, we present a computationally tractable contact model in~\prettyref{sec:slow}, which satisfies all these properties as proved in our~\prettyref{appen:proof} and is applicable to arbitrary articulated bodies represented using triangle meshes. Further in~\prettyref{sec:fast}, we significantly improve the computational efficiency for evaluating the contact model using a Bounding Sphere Hierarchy (BSH). 
We have incorporated our contact model into a full-featured rigid body simulator and experimented on a row of robotic manipulation and locomotion tasks. Our results show that our method can discover complex, contact-rich control signals from trivial initialization, while previous models can get optimizers stuck at trivial local minima or suffer from slow convergence.
\section{Related Work}
The idea of differentiable physics originates from the pioneering work~\cite{5979814}, which is then built into the MuJoCo simulator~\citep{6386025}, where gradients are computed via costly finite difference schemes. Differentiable simulators are then proposed to use more efficient analytic gradient information. Early works apply this idea to rigid body simulators~\citep{de2018end} and reduced-order deformable body simulators~\citep{pan2018active}. The idea is then adopted in other simulator models~\citep{newbury2024review} for elastic and plastic deformations, articulated bodies, and fluid bodies, to name just a few. Since their invention, differentiable simulators have found many applications in computer graphics, robotics, and machine learning. Early works along this line use differentiable simulators to perform model-predictive control~\citep{6386025} and guide deep policy search~\citep{levine2013guided}. Differentiable simulators can be combined with appearance models to perform state estimation~\citep{ma2022risp} and system identification~\citep{le2021differentiable}. In computer graphics, animators use differentiable simulators to inversely design initial and boundary conditions~\citep{li2023difffr,stuyck2023diffxpbd,Du2021diffpd}. They can also provide gradient information for physics-informed machine learning such as neural PDE solvers~\citep{heiden2021neuralsim} and neural motion planning~\citep{ijcai2019p869}.

Certain substeps in a simulation procedure are inherently non-differentiable, of which the most important substep is contact handling. It is known that gradient information is lost for collision detection with thin-shell-like objects~\citep{harmon2009asynchronous,li2022contact, Li2022DiffCloth} and the sudden change of contact forces in collision responses incur non-smoothness, which requires manually choosing a specific direction in the Clark subdifferential~\citep{Werling-RSS-21}. These factors can compromise the gradient information, hindering downstream optimizers' performance. In~\cite{suh2022differentiable}, authors propose a mixed gradient-free and gradient-based optimizer to boost the performance of policy search. On a parallel front,~\cite{li2020incremental} showed that the contact model can be reformulated to prevent gradient vanishing for thin-shell-like objects. Differentiable simulator~\citep{huang2024differentiable} built on top of this technique exhibits better robustness. However, we show that even with this technique, optimizers can still suffer from vanishing gradients.
\section{\label{sec:property}Differentiable Physics with Well-behaved Contact Model}
In this section, we first formulate the problem of a differentiable physical model, and then formalize the properties of a well-behaved contact model that provides useful gradient information. 

\subsection{Differentiable Physics Model}
Throughout the paper, we consider articulated bodies geometrically represented using triangle meshes. This is the representation adopted by a majority of differentiable contact models~\citep{Werling-RSS-21,huang2024differentiable,Xu-RSS-21}. Formally, we assume the configuration of an articulated body is represented using a set of $V$ vertices located at $x_{1,\cdots,V}\in\mathbb{R}^3$, and we use $x$ without subscript to denote the concatenated vertex vector $x\in\mathbb{R}^{3V}$. The vertices are connected to form a set of $T$ triangles $t_{1,\cdots,T}$, with each cornering three vertices and defined as $t_i=\{i(1),i(2),i(3)\}\subset\{1,\cdots,V\}$. The configuration space $x\in\CAll\subset\mathbb{R}^{3V}$ can be divided into a penetrating set $\CObs=\{x\in\CAll|\exists t_i\neq t_j \land \CH(x_{i(k)\in t_i})\cap\CH(x_{j(k)\in t_j})\neq\emptyset\}$, where $\CH$ is the closed convex hull of a set of vertices, and a penetration-free set $\CFree=\CAll/\CObs$. The goal of a contact model is to impose contact forces on the rigid object to ensure that $x\in\CFree$.

A discrete-time physical model can be cast as a time transition function $x^{t+1}=f(x^t,x^{t-1},\delta t)$, where we use superscript to denote the time index, $x^t$ is the kinematic state of a body at the $t$th time instance, and $\delta t$ is the timestep size. The concatenation of two kinematic states $\langle x^t,x^{t-1}\rangle$ composes a dynamic state of the body, with velocity approximated as $(x^t-x^{t-1})/\delta t$. It is well-known~\citep{marsden2001discrete,gast2015optimization} that the transition function can be cast as a numerical optimization:
\begin{align}
\label{eq:opt}
x^{t+1}\in\argmin{}_{x_\star^{t+1}}\mathcal{L}(x_\star^{t+1},x^t,x^{t-1},\delta t),
\end{align}
which leads to stable performance of modern differentiable position-based dynamics, such as~\cite{huang2024differentiable}. In these position-based models, the Lagrangian function $\mathcal{L}$ contains various terms that model different behaviors. Specifically, we define:
\begin{align*}
\mathcal{L}(x_\star^{t+1},x^t,x^{t-1},\delta t)=
\mathcal{I}(x_\star^{t+1},x^t,x^{t-1},\delta t)+
\mu\mathcal{P}(x_\star^{t+1})+
\mathcal{D}(x_\star^{t+1},x^t,\delta t),
\end{align*}
where the term $\mathcal{I}$ models inertial acceleration, $\mathcal{P}$ models the contact potential weighted by a parameter $\mu$, and $\mathcal{D}$ models frictional damping. We refer readers to~\cite{huang2024differentiable} for more details of other terms and we focus on the term $\mathcal{P}$ in this paper. In optimization-based simulators, the contact mechanics are mainly determined by the potential $\mathcal{P}$, whose proper definition has been discussed, e.g., in~\cite{fisher2001deformed,harmon2009asynchronous,li2020incremental,ye2024sdrs}.

\subsection{Properties of Contact Potential}
\begin{figure*}[ht]
\centering
\includegraphics[width=.75\linewidth]{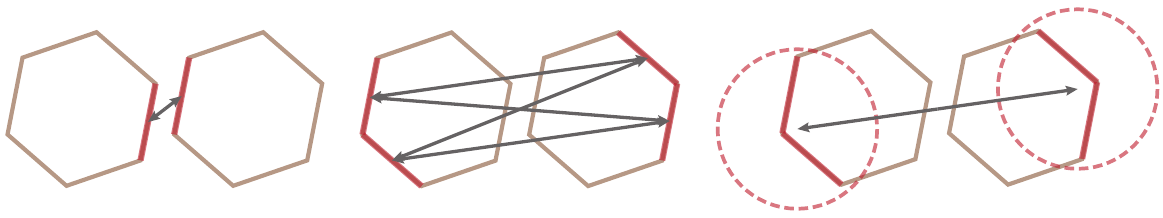}
\caption{\label{fig:propertyC}\footnotesize{A 2D illustration of our contact force between a pair of hexagons satisfying~\prettyref{prop:C}. Each pentagons have 5 line segments in 2D (resp. triangles in 3D). Left: We compute the exact contact force (arrow) between each pair of nearby line segments. Middle: For faraway pairs of line segments, computing exact contact forces would involve too many segment pairs, e.g. 4 pairs of forces between 2 edges on each pentagon. Right: Instead, we group faraway segments and approximate the contact forces between centers of bounding circles in 2D (resp. bounding spheres in 3D).}}
\end{figure*}

\begin{figure*}[ht]
\centering
\scalebox{.75}{
\setlength{\tabcolsep}{1px}
\begin{tabular}{ccccc}
\toprule
\multirow{2}{*}[-25px]{\rotatebox[origin=c]{90}{Failure Case}}&
Log-barrier (\ref{prop:A})&
Smoothness (\ref{prop:B})&
Non-prehensile (\ref{prop:C}a)&
Non-vanishing (\ref{prop:C}b)\\
&
\includegraphics[width=.23\textwidth]{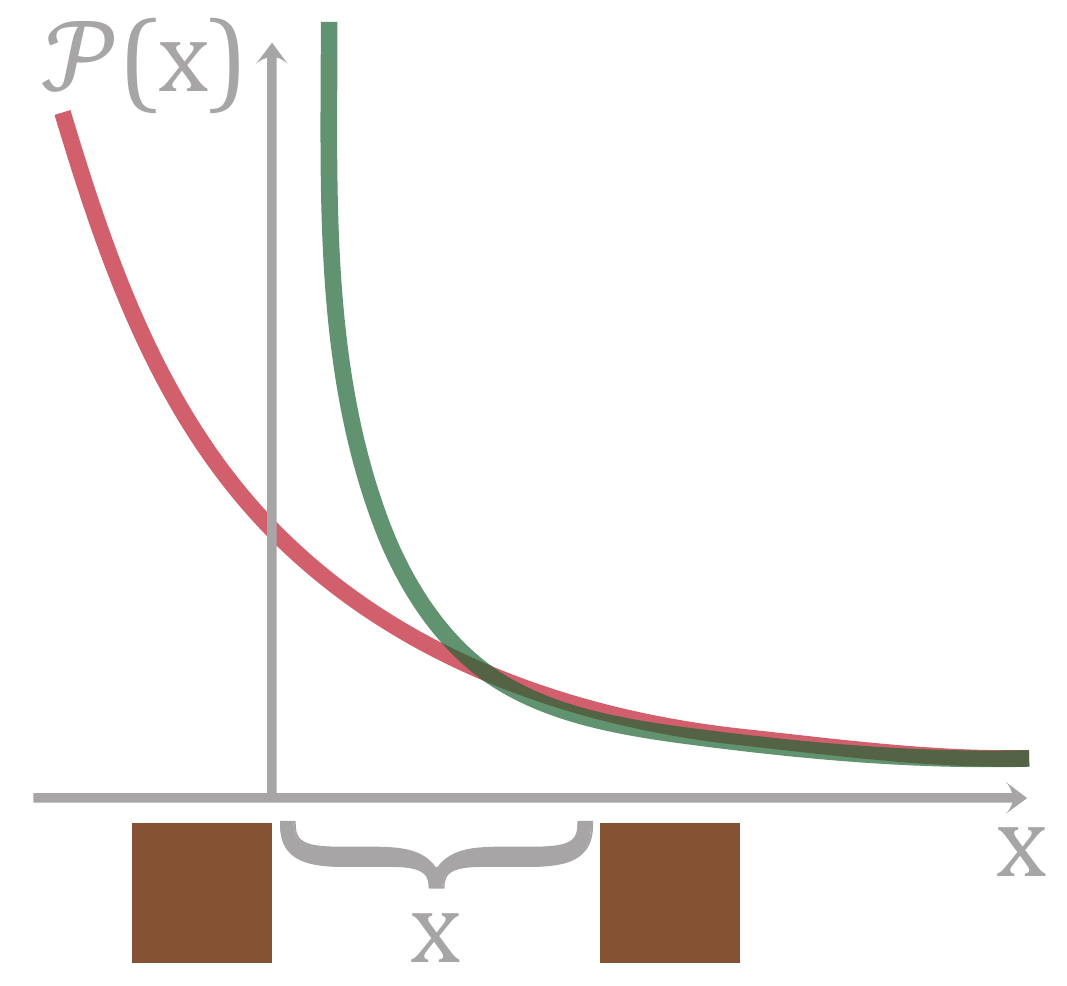}& \includegraphics[width=.23\textwidth]{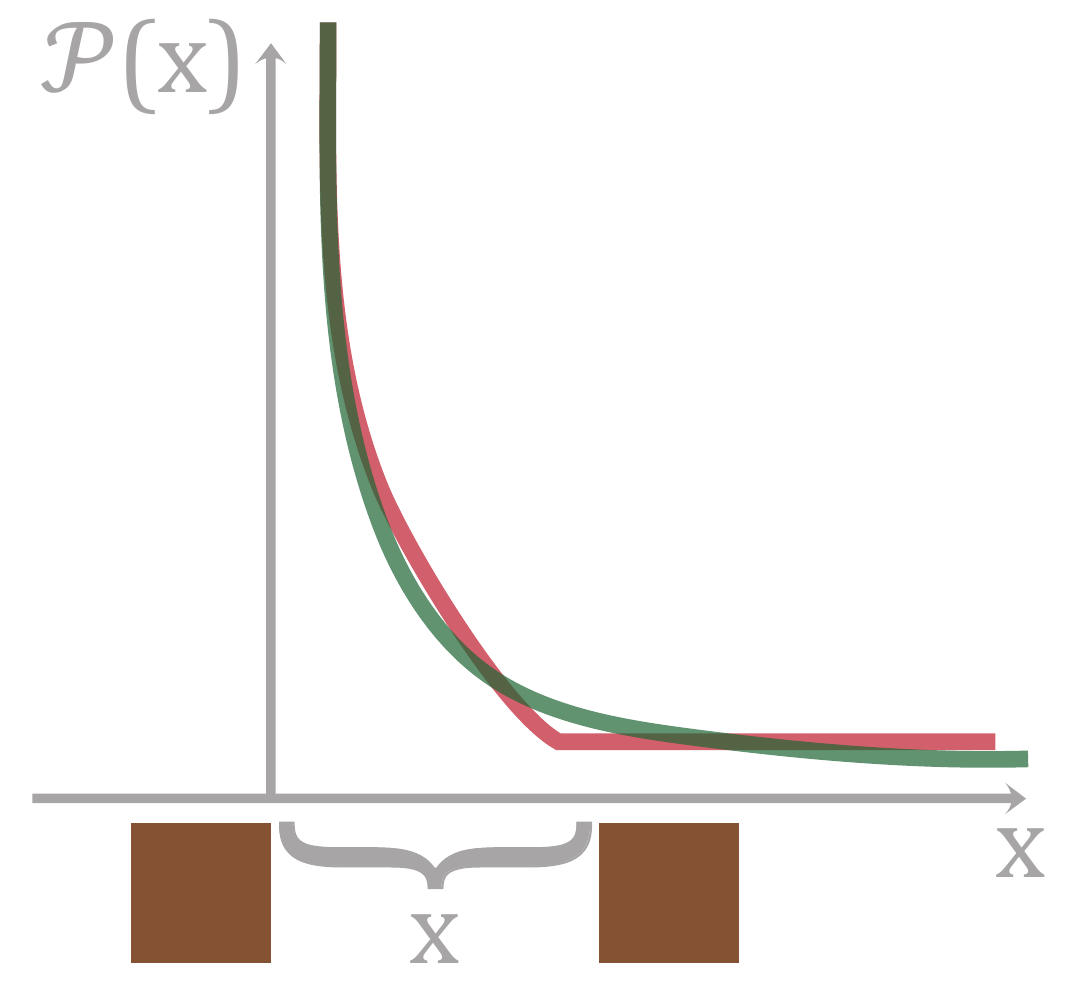}& \includegraphics[width=.23\textwidth]{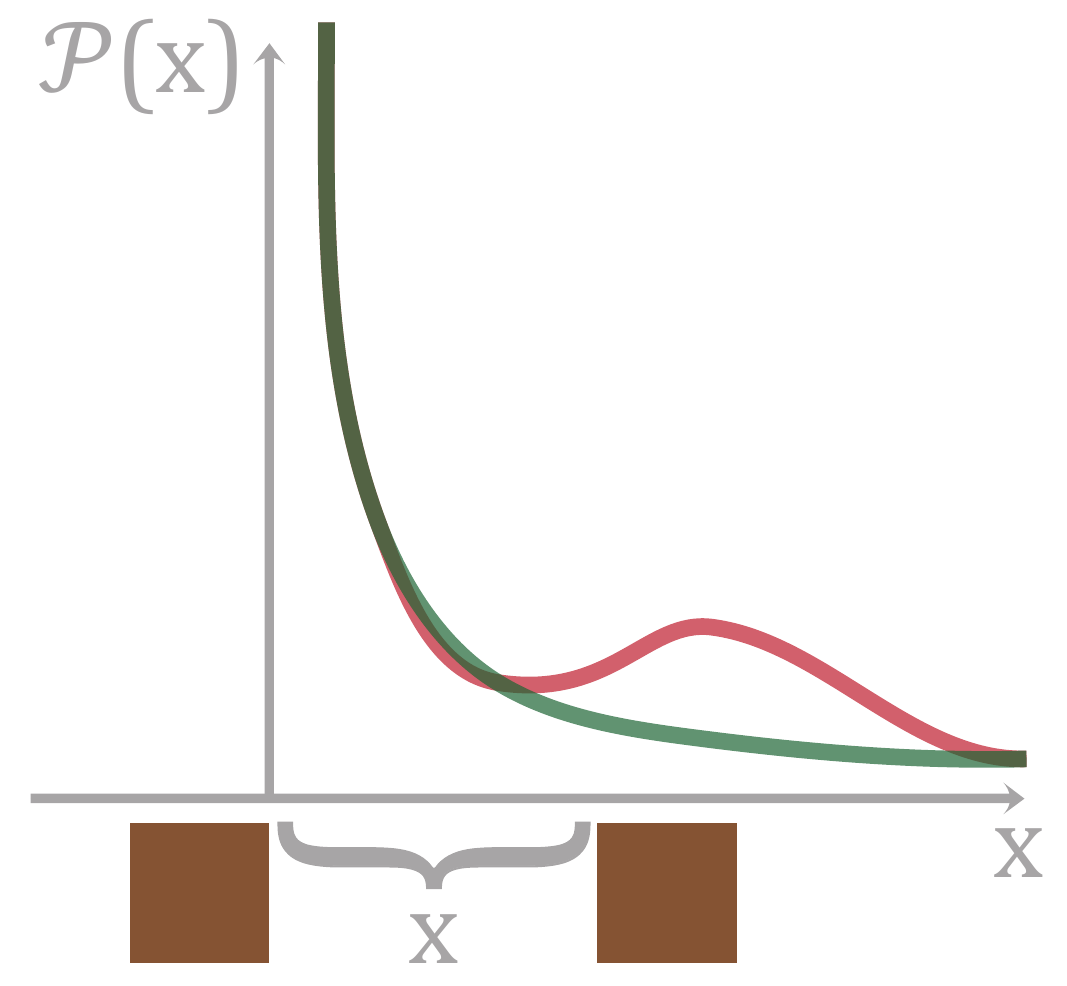}& \includegraphics[width=.23\textwidth]{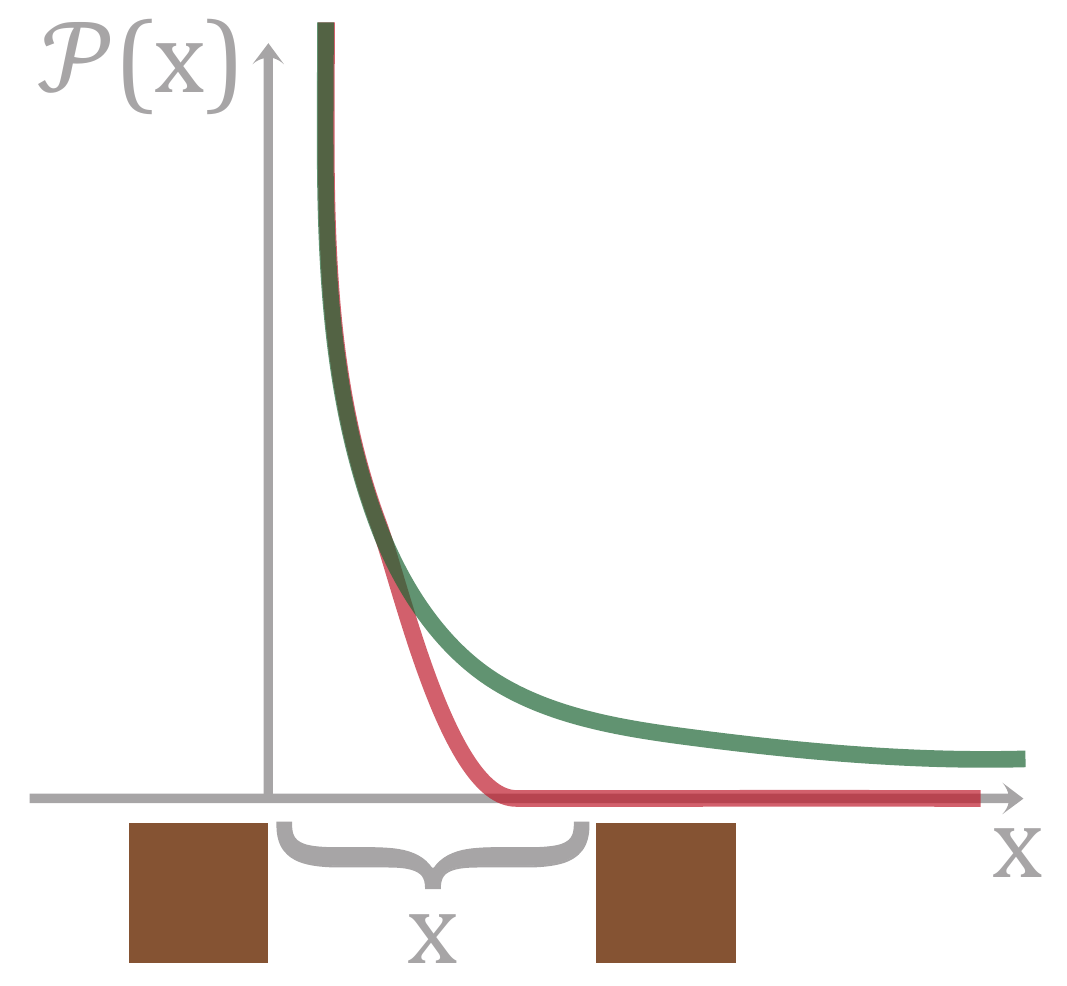}\\
\bottomrule
\end{tabular}}
\caption{\label{fig:failure}\footnotesize{We illustrate failure cases for various properties, where we assume two boxes (brown) are separated by distance $x$ and plot the contact potential $\mathcal{P}(x)$ that pertains the property in green and fails the property in red. Our~\prettyref{prop:A} requires $\mathcal{P}$ to tend to infinity as $x\to0^+$. \prettyref{prop:B} requires $\mathcal{P}$ to have well-defined second-order derivatives. \prettyref{prop:C}a requires the contact force to always push the two boxes apart. \prettyref{prop:C}b requires $\mathcal{P}$ and thus the contact force to be non-vanishing for arbitrarily large $x$.}}
\vspace{-10px}
\end{figure*}

We propose four aspects of indispensable properties of a well-behaved contact potential. \textbf{Log-barrier:} It is well-known in the theory of continuous collision handling~\citep{brochu2012efficient} that a penetrating state $x^t\in\CObs$ corresponds to non-smooth landscape of the contact potential. Therefore, a well-behaved contact potential should provably prevent any penetrations by ensuring that $x^t\in\CFree$ for any $t$. This property is first proposed and achieved in~\cite{harmon2009asynchronous}, where authors designed $\mathcal{P}$ to be a positive layered potential function that is infinite when $x^t\in\CObs$. As a result, a numerical optimizer with globalization techniques, such as line search and trust-region, can ensure the monotonic decrease of the Lagrangian $\mathcal{L}$, leading to the finiteness of $\mathcal{P}$ and thus $x^{t+1}\in\CFree$. \cite{li2020incremental} further elucidates that a potential $\mathcal{P}$ should act as a log-barrier function in interior point optimization, which is formalized as our first property below:
\begin{property}[Log-barrier]
\label{prop:A}
$\mathcal{P}(x)\geq0$ is continuous for any $x\in\CAll$ and $\mathcal{P}(x)=\infty$ iff $x\in\CObs$.
\end{property}
Note that~\prettyref{prop:A} does not describe the exact contact mechanics, because an exact contact model can only impose contact forces on bodies when they are exactly touching, i.e., $x\in\partial\CObs$, but our contact potential induces the generalized contact force $-\FPPR{\mathcal{P}}{x}$ even when $x\notin\CObs$. To mitigate this issue,~\cite{li2020incremental} proposes to iteratively approximate the true contact mechanics by tuning the coefficient $\mu$. Indeed, we can easily verify that $\lim_{\mu\to0^+}\mu\mathcal{P}$ converges to the indicator function that equals to $\infty$ if $x\in\partial\CObs$ and 0 otherwise. 
\textbf{Smoothness:} In order to utilize the primal log-barrier method for computing $x^{t+1}$ by solving optimization would require $\mathcal{P}$ to be at least differentiable. To this end,~\cite{li2020incremental} proposed a differentiable surrogate of the triangle-triangle distance function. Unfortunately, although differentiability is enough for solving~\prettyref{eq:opt}, it is not enough for providing reliable gradient information. Indeed, the gradient of a numerical optimization takes the following form by the implicit function theorem:
\begin{align*}
\FPP{x^{t+1}}{\TWO{x^t}{x^{t-1}}}=-\left[\FPPT{\mathcal{L}}{x^{t+1}}\right]^{-1}\FPPTT{\mathcal{L}}{x^{t+1}}{\TWO{x^t}{x^{t-1}}},
\end{align*}
whose proper evaluation requires $\mathcal{P}$ to be twice-differentiable, which is not satisfied in~\cite{li2020incremental,huang2024differentiable} as shown in our~\prettyref{appen:IPC}, leading to our second property:
\begin{property}[Smoothness]
\label{prop:B}
$\mathcal{P}$ is twice differentiable at $x\in\CFree$.
\end{property}
Unfortunately, being numerically well-defined does not guarantee that the gradient information can effectively guide the optimizer to find meaningful solutions for downstream applications. To this end, we introduce two other properties that ensure the contact model is non-prehensile and non-vanishing. \textbf{Non-prehensile:} We know that a passive contact can only impose unilateral pushing forces between a pair of contacting objects, instead of pulling objects together. Formally, we introduce a sufficient condition to ensure non-prehensile forces. Let us define two index subsets of well-separated vertices $\mathcal{I}\cap\mathcal{J}=\emptyset$ and $\mathcal{I}\cup\mathcal{J}\subseteq\{1,\cdots,V\}$, such that the convex hull of these sets of vertices are non-overlapping, i.e. $\CH(x_{i\in\mathcal{I}})\cap\CH(x_{j\in\mathcal{J}})=\emptyset$. A contact potential between these two subsets can be defined as a pair-wise contact term
$\mathcal{P}^{\mathcal{I}\cup\mathcal{J}}(x_{i\in\mathcal{I}},x_{j\in\mathcal{J}})$ or $\mathcal{P}^{\mathcal{I}\cup\mathcal{J}}$ for short. Note that we can also establish~\prettyref{prop:A} and~\prettyref{prop:B} for a pair-wise contact term $\mathcal{P}^{\mathcal{I}\cup\mathcal{J}}$. To this end, we define $\CObs^{\mathcal{I}\cup\mathcal{J}}=\{x\in\mathcal{C}|\exists t_i\neq t_j\land t_i\cup t_j\subseteq\mathcal{I}\cup\mathcal{J}, \CH(x_{i'\in t_i})\cap\CH(x_{j'\in t_j})\neq\emptyset\}$ and $\CFree^{\mathcal{I}\cup\mathcal{J}}=\CAll/\CObs^{\mathcal{I}\cup\mathcal{J}}$ and we have the following~\prettyref{prop:A} and~\prettyref{prop:B} for pairwise contact terms:
\begin{define}
$\mathcal{P}^{\mathcal{I}\cup\mathcal{J}}$ pertains~\prettyref{prop:A} if $\mathcal{P}^{\mathcal{I}\cup\mathcal{J}}\geq0$ for any $x\in\CAll$ and $\mathcal{P}^{\mathcal{I}\cup\mathcal{J}}=\infty$ iff $x\in\CObs^{\mathcal{I}\cup\mathcal{J}}$. $\mathcal{P}^{\mathcal{I}\cup\mathcal{J}}$ pertains~\prettyref{prop:B} if it is twice differentiable at $x\in\CFree^{\mathcal{I}\cup\mathcal{J}}$.
\end{define}
$\mathcal{P}^{\mathcal{I}\cup\mathcal{J}}$ induces the following contact force on any $x_{i\in\mathcal{I}}$ or $x_{j\in\mathcal{J}}$:
\begin{align*}
f_{i\in\mathcal{I}}^{\mathcal{I}\cup\mathcal{J}}=-\FPP{}{x_{i\in\mathcal{I}}}\mathcal{P}^{\mathcal{I}\cup\mathcal{J}}(x_{i\in\mathcal{I}},x_{j\in\mathcal{J}})\quad
f_{j\in\mathcal{J}}^{\mathcal{I}\cup\mathcal{J}}=-\FPP{}{x_{j\in\mathcal{J}}}\mathcal{P}^{\mathcal{I}\cup\mathcal{J}}(x_{i\in\mathcal{I}},x_{j\in\mathcal{J}}).
\end{align*}
To allow only non-prehensile forces, we require that each $f_i^{\mathcal{I}\cup\mathcal{J}}$ is pointing from $\CH(x_{j\in\mathcal{J}})$ to $\CH(x_{i\in\mathcal{I}})$ and vice versa. Formally, we define the set of non-zero vectors pointing from $\CH(x_{j\in\mathcal{J}})$ to $\CH(x_{i\in\mathcal{I}})$ as
$\mathcal{F}_{\mathcal{J}\to\mathcal{I}}=\{\alpha(a-b)|\alpha>0\land
a\in\CH(x_{i\in\mathcal{I}})\land
b\in\CH(x_{j\in\mathcal{J}})\}$, and require that:
\begin{align}
\label{eq:non-prehensile}
\forall i\in\mathcal{I}: f_i^{\mathcal{I}\cup\mathcal{J}}\in
\mathcal{F}_{\mathcal{J}\to\mathcal{I}}\text{ and }
\forall j\in\mathcal{J}: f_j^{\mathcal{I}\cup\mathcal{J}}\in
\mathcal{F}_{\mathcal{I}\to\mathcal{J}}.
\end{align}
\textbf{Non-vanishing:} Our final property is of paramount importance and ensures that a differentiable simulator provides non-zero gradient information at arbitrary configuration. This is ensured by our definition of the non-prehensile force set $\mathcal{F}_{\mathcal{J}\to\mathcal{I}}$. Indeed, since $\CH(x_{j\in\mathcal{J}})$ and $\CH(x_{i\in\mathcal{I}})$ are disjoint, closed convex sets, for any $\alpha(a-b)\in\mathcal{F}_{\mathcal{J}\to\mathcal{I}}$, we have $a\neq b$ and $\alpha>0$, leading to $f_i^{\mathcal{I}\cup\mathcal{J}}\neq0$ for all $i\in\mathcal{I}$. We further ensure that the contact forces between every pair of geometric primitives (triangles) are taken into consideration. Put together, we can ensure both properties by requiring that $\mathcal{P}$ is a summation of pairwise contact terms $\mathcal{P}^{\mathcal{I}\cup\mathcal{J}}$ between well-separated vertex clusters:
\begin{property}[Non-prehensile \& Non-vanishing]
\label{prop:C}
At every $x\in\CFree$, we can define a finite family of set pairs $\mathcal{A}(x)=\{\langle \mathcal{I},\mathcal{J}\rangle|\mathcal{I}\cap\mathcal{J}=\emptyset\land\mathcal{I}\cup\mathcal{J}\subseteq\{1,\cdots,V\}\land\CH(x_{i\in\mathcal{I}})\cap\CH(x_{j\in\mathcal{J}})=\emptyset\}$. We have
$\mathcal{P}=\sum_{\langle \mathcal{I},\mathcal{J}\rangle\in\mathcal{A}(x)}\mathcal{P}^{\mathcal{I}\cup\mathcal{J}}$ such that every term $\mathcal{P}^{\mathcal{I}\cup\mathcal{J}}$ satisfy~\prettyref{eq:non-prehensile}, and for every pair of triangles $\langle t_i,t_j\rangle$ on different rigid bodies, we have $t_i\cup t_j\subseteq\mathcal{I}\cup\mathcal{J}$ for at least one $\langle \mathcal{I},\mathcal{J}\rangle\in\mathcal{A}(x)$. 
\end{property}
This is an important property that allows the gradient information to be provided for arbitrarily distant objects. In many applications, such gradient information can help a local optimizer discover contact-rich motions from trivial initial guesses. Regretfully, we are not aware of any contact model that pertains~\prettyref{prop:A},\ref{prop:B},\ref{prop:C} at the same time. We summarize the failure cases for various properties in~\prettyref{fig:failure} and compare the property completeness in~\prettyref{table:failure}. In~\prettyref{fig:propertyC}, we illustrate the main idea behind our contact model in~\prettyref{sec:fast} that satisfies~\prettyref{prop:C}.
\begin{table*}[ht]
\centering
\scalebox{.75}{
\setlength{\tabcolsep}{10px}
\begin{tabular}{ccccc}
\toprule
Formulation&
\ref{prop:A}&
\ref{prop:B}&
\ref{prop:C}a&
\ref{prop:C}b\\
\midrule
\cite{Werling-RSS-21,xu2022accelerated}
&\xmark&\xmark&\cmark&\xmark\\
\midrule
\cite{fisher2001deformed,guendelman2003nonconvex}
&\xmark&\xmark&\cmark&\xmark\\
\midrule
\cite{harmon2009asynchronous,li2020incremental}
&\cmark&\xmark&\cmark&\xmark\\
\midrule
\cite{ye2024sdrs}
&\cmark&\cmark&\cmark&\xmark\\
\midrule
Ours
&\cmark&\cmark&\cmark&\cmark\\
\bottomrule
\end{tabular}}
\caption{\label{table:failure}\footnotesize{Comparison of property completeness (\ref{prop:A}: Log-barrier, \ref{prop:B}: Smoothness, \ref{prop:C}a: Non-prehensile, \ref{prop:C}b: Non-vanishing). Contact models based on complementary conditions~\cite{Werling-RSS-21,xu2022accelerated} or soft penalty functions~\cite{fisher2001deformed,guendelman2003nonconvex} cannot guarantee intersection-free or sufficient smoothness. Contact models based on the log-barrier functions~\cite{harmon2009asynchronous,li2020incremental} only have first-order derivatives, which does not support differentiation using the inverse function theorem. Finally, prior contact models have vanishing gradient when the distance is larger than a small margin.}}
\vspace{-10px}
\end{table*}
\section{\label{sec:slow}Well-behaved Contact Potential}
In this section, we propose a practical and well-behaved contact potential. We start by showing that slightly modifying an existing contact potential~\citep{liang2024second,ye2024sdrs} makes it well-behaved, but such a potential is slow to compute. We then improve its computational efficacy in~\prettyref{sec:fast} by borrowing ideas from the well-known hierarchical algorithm~\citep{barnes1986hierarchical} for N-body simulation.

Our~\prettyref{prop:A} requires that our contact potential acts as a primal barrier function. However, existing barrier potential functions~\citep{harmon2009asynchronous,li2020incremental} is derived from a modified triangle-triangle distance function, denoted as $d(x_{i(k)\in t_i},x_{j(k)\in t_j})$, which is then assembled to form the following contact potential: $\mathcal{P}=\sum_{t_i\neq t_j}P(d(x_{i(k)\in t_i},x_{j(k)\in t_j}))$, with $P$ being some locally supported barrier function. Unfortunately, this potential is at most first-order differentiable and violates~\prettyref{prop:B} and the local support of $P$ violates~\prettyref{prop:C}. Instead, we propose to adopt the more general contact potential~\citep{liang2024second,ye2024sdrs} between a pair of convex hulls. Since a triangle is convex by nature, the more general contact potential can be adopted to serve as $\mathcal{P}^{t_i\cup t_j}$. Specifically, given a pair of triangles $t_i$ and $t_j$, since the two triangles are both convex sets, we can define a separating plane $p_{ij}=\TWO{n_{ij}^T}{d_{ij}}^T\in\mathbb{R}^4$ between them if the two sets are disjoint, with $n_{ij}$ and $d_{ij}$ being the normal and offset, such that: $\langle x_{i(k)\in t_i},n_{ij}\rangle+d_{ij}>0$ and $\langle x_{j(k)\in t_j},n_{ij}\rangle+d_{ij}<0$. As a result, we introduce the following potential via a nested optimization:
\footnotesize
\begin{align*}
\mathcal{P}^{t_i\cup t_j}=&\min{p_{ij}}\mathcal{L}_{ij}(p_{ij},x_{i(k)\in t_i},x_{j(k)\in t_j})\\
\mathcal{L}_{ij}(p_{ij},x_{i(k)\in t_i},x_{j(k)\in t_j})=&\left[12\PXInv{1-\|n_{ij}\|}+
\sum_{k=1}^3\PXInv{\langle x_{i(k)},n_{ij}\rangle+d_{ij}}+
\sum_{k=1}^3\PXInv{\langle x_{j(k)},-n_{ij}\rangle-d_{ij}}\right],
\end{align*}
\normalsize
where we purposefully introduce a constant coefficient $12$ in the first term so that our follow-up derivations take a simpler form, and other positive coefficients can be used as well. As a main point of departure from the original formulation in~\cite{liang2024second,ye2024sdrs}, we do not use locally supported log-barrier function, but define the potential function $1/(\bullet)^+=1/\max(\bullet,0)$ that has a global support on $\mathbb{R}^+$ to prevent vanishing gradient. Note that $n_{ij}, d_{ij}$ are computed by minimizing $\mathcal{L}_{ij}$ and thus $n_{ij}$ is not normalized. It is easy to verify that the objective function defined in $\mathcal{P}^{t_i\cup t_j}$ is a strictly convex function with a unique minimizer, so that $\mathcal{P}^{t_i\cup t_j}$ is a well-defined function. With the pair-wise potential defined, we can assemble them and define:
\begin{align}
\label{eq:potential-slow}
\mathcal{P}=\sum_{t_i\neq t_j}\mathcal{P}^{t_i\cup t_j}(x_{i(k)\in t_i},x_{j(k)\in t_j}),
\end{align}
where the summation is taken over triangle pairs on different rigid bodies. We now show that the so-defined contact potential pertains all our desired properties.
\begin{lemma}
\label{lem:slow}
Each pair-wise potential $\mathcal{P}^{t_i\cup t_j}$ in \prettyref{eq:potential-slow} pertains~\prettyref{prop:A}, \prettyref{prop:B}, and satisfies~\prettyref{eq:non-prehensile}, so that the potential $\mathcal{P}$ pertains~\prettyref{prop:A},\ref{prop:B},\ref{prop:C}.
\end{lemma}
At this point, we have shown that~\prettyref{eq:potential-slow} is a well-behaved contact potential function. Remarkably, this function is computationally practical. Indeed, each term $\mathcal{P}^{t_i\cup t_j}$ involves a small 4D-optimization problem with a strictly convex objective function, which can be solved efficiently using Newton's method to evaluate $p_{ij}$. The first and second derivatives of $\mathcal{P}^{t_i\cup t_j}$ can then be computed using the inverse function theorem. However, a brute force computation of the potential function $\mathcal{P}$ is not efficient, since it involves terms 
that account for the contact potential between each pair of disjoint triangles which increases in the square order of triangles.
\section{\label{sec:fast}Efficient Contact Potential Evaluation}
The computational challenge of evaluating~\prettyref{eq:potential-slow} lies in accounting for the contact potentials between all pairs of disjoint triangles. This scenario closely parallels the N-body simulation problem, where the forces between all pairs of particles must be computed. Instead of performing a brute-force summation with a computational cost of $O(N^2)$, efficient algorithms such as the tree code~\citep{barnes1986hierarchical} and the fast multipole expansion~\citep{greengard1987fast} achieve costs of $O(N\log(N))$ and $O(N)$, respectively. These methods rely on the multipole expansion to separate the influences of source particles from those of target particles.
However, several factors make these algorithms unsuitable for our case. First, the multipole expansions for our contact potential $\mathcal{P}^{t_i\cup t_j}$ remain undefined. Second, even if such expansions could be derived, the abrupt transition between the exact potential and its multipole approximation could introduce discontinuities, thereby violating~\prettyref{prop:B}. Inspired by the fast multipole method~\citep{greengard1987fast}, we propose instead a modified potential that is also well-behaved and can be evaluated hierarchically. Our main idea is to smoothly transit from the exact potential function $\mathcal{P}^{t_i\cup t_j}$ to simplified functions that can be hierarchically evaluated. 

\subsection{Smooth Transition Between Potentials}
Let us consider the pairwise potential between index set $\mathcal{I}$ and $\mathcal{J}$. For an index set $\mathcal{I}$, we define its bounding sphere to be centered at $x_\mathcal{I}=\sum_{i\in\mathcal{I}}x_i/|\mathcal{I}|$ with radius $R_\mathcal{I}\geq\max_{i\in\mathcal{I}}|x_\mathcal{I}-x_i|$. Suppose there are two versions of the potential denoted as $\mathcal{P}_{d_1}^{\mathcal{I}\cup\mathcal{J}}$ and $\mathcal{P}_{d_2}^{\mathcal{I}\cup\mathcal{J}}$, we can smoothly blend the two functions when the distance between $x_\mathcal{I}$ and $x_{\mathcal{J}}$ grows from $d_1$ to $d_2$ with $d_1<d_2$ as illustrated in~\prettyref{fig:BSH-idea}, yielding the following blending potential:
\begin{align}
\label{eq:blending}
\mathcal{P}_{d_1\to d_2}^{\mathcal{I}\cup\mathcal{J}}=
(1-\phi_{d_1\to d_2}(x))\mathcal{P}_{d_1}^{\mathcal{I}\cup\mathcal{J}}+\phi_{d_1\to d_2}(x)\mathcal{P}_{d_2}^{\mathcal{I}\cup\mathcal{J}},
\end{align}
where we define the interpolation function as:
\begin{align*}
\phi_{d_1\to d_2}(x)=\Phi((\|x_{\mathcal{I}}-x_{\mathcal{J}}\|-d_1)/(d_2-d_1))\text{ and }
\Phi(d)=\max(\min(6d^5-15d^4+10d^3,1),0).
\end{align*}
Similar to the tree code algorithm~\citep{barnes1986hierarchical}, our goal of blending is to gradually replace exact potential functions with faster-to-compute approximations. Such blending should not happen when two sets of vertices are too close to each other. 
In practice, we only allow blending when $R_\mathcal{I}+R_\mathcal{J}\leq d_1$ where $R_\mathcal{I},R_\mathcal{J}$ are the radii of the bounding spheres of $\mathcal{I},\mathcal{J}$.
We first show the well-behaved nature of potential functions is invariant to blending:
\begin{lemma}
\label{lem:blending}
Taking the following assumptions: i) $R_\mathcal{I}+R_\mathcal{J}\leq d_1<d_2$; ii) $\mathcal{P}_{d_1}^{\mathcal{I}\cup\mathcal{J}}$ pertains~\prettyref{prop:A},~\prettyref{prop:B}, and satisfies~\prettyref{eq:non-prehensile}; iii) $0\leq\mathcal{P}_{d_2}^{\mathcal{I}\cup\mathcal{J}}\leq\mathcal{P}_{d_1}^{\mathcal{I}\cup\mathcal{J}}$ when $\|x_\mathcal{I}-x_\mathcal{J}\|\geq d_1$; iv) $\mathcal{P}_{d_2}^{\mathcal{I}\cup\mathcal{J}}$ has~\prettyref{prop:B}, and satisfies~\prettyref{eq:non-prehensile},
then $\mathcal{P}_{d_1\to d_2}^{\mathcal{I}\cup\mathcal{J}}$ has the same properties as $\mathcal{P}_{d_1}^{\mathcal{I}\cup\mathcal{J}}$.
\end{lemma}
\prettyref{lem:blending} can be immediately used to blend our potential function $\mathcal{P}^{t_i\cup t_j}$ with a much simpler, closed-form function. Consider moving all three vertices of $t_i$ to the center point $x_{t_i}=(x_{i(1)}+x_{i(2)}+x_{i(3)})/3$ and similarly moving $t_j$ to $x_{t_j}$, then the potential $\mathcal{P}^{t_i\cup t_j}$ takes the following (centered) form after some basic algebraic manipulation:
\begin{equation}
\label{eq:centered-potential}
\ResizedEq{\mathcal{P}_c^{t_i\cup t_j}=&
\argmin{p_{ij}}\left[12\PXInv{1-\|n_{ij}\|}+
\sum_{k=1}^3\PXInv{\langle x_{t_i},n_{ij}\rangle+d_{ij}}+
\sum_{k=1}^3\PXInv{\langle x_{t_j},-n_{ij}\rangle-d_{ij}}\right]
=12\left[1+\frac{1}{\|x_{t_i}-x_{t_j}\|^{1/2}}\right]^2.}
\end{equation}
We are now ready to apply~\prettyref{lem:blending} to blend $\mathcal{P}^{t_i\cup t_j}$ and $\mathcal{P}_c^{t_i\cup t_j}$ in a well-behaved manner:
\begin{corollary}
\label{cor:blending}
If we define $\mathcal{P}_{d_1}^{t_i\cup t_j}=\mathcal{P}^{t_i\cup t_j}$ and $\mathcal{P}_{d_2}^{t_i\cup t_j}=\mathcal{P}_c^{t_i\cup t_j}$, then $\mathcal{P}_{d_1\to d_2}^{t_i\cup t_j}$ pertains~\prettyref{prop:A},~\prettyref{prop:B}, and satisfies~\prettyref{eq:non-prehensile}.
\end{corollary}
Intuitively, \prettyref{cor:blending} allows us to use the exact potential $\mathcal{P}^{t_i\cup t_j}$ when two triangles are very close to each other, while switching to a simpler, closed-form potential $\mathcal{P}_c^{t_i\cup t_j}$ when the centers of two triangles are well-separated by some distance $d_2$.

\subsection{Hierarchical Potential Blending}
\begin{wrapfigure}{r}{0.5\textwidth}
\centering
\vspace{-40px}
\includegraphics[width=0.48\textwidth]{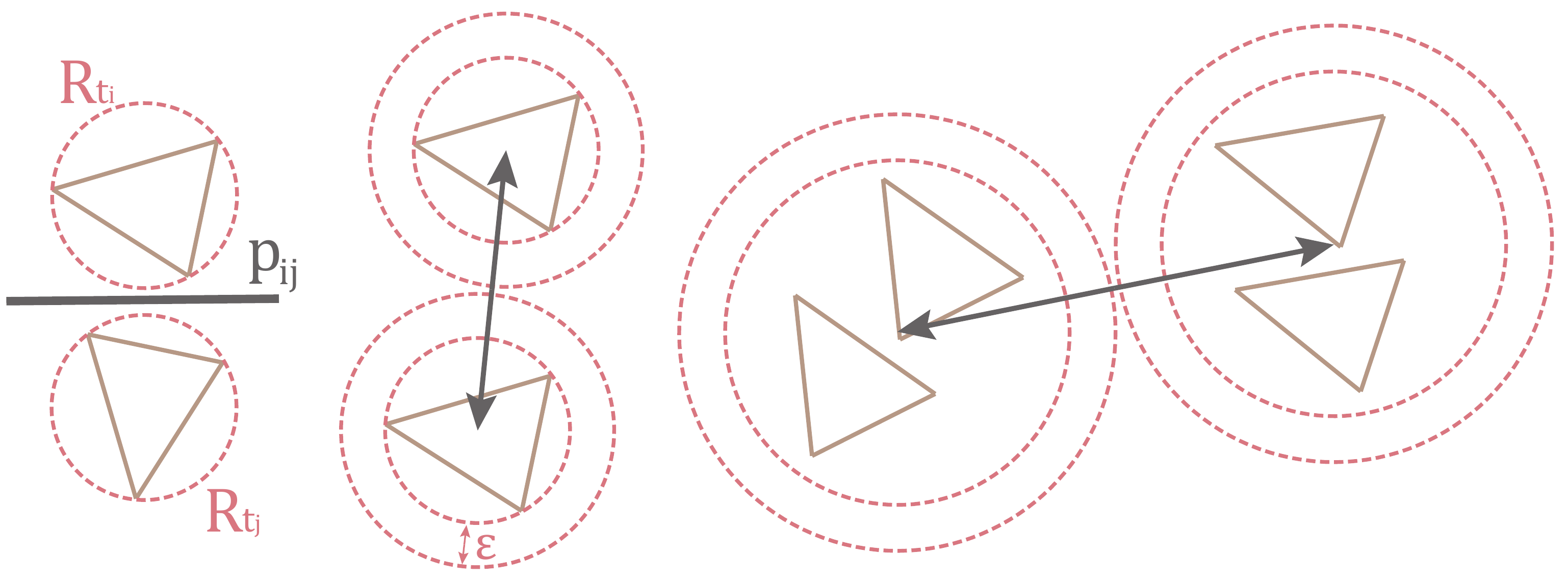}
\caption{\label{fig:BSH-idea}\footnotesize{Illustration of our BSH-based contact potential. When two triangles are nearby, we use the exact potential based on separating plane $p_{ij}$ (left). When the center of bounding sphere is separated by at least $(R_{t_i}+R_{t_j})(1+\epsilon)$, we use the centered potential in~\prettyref{eq:centered-cluster-potential} (middle). These two cases are combined by smooth blending. The centered potential can be calculated hierarchically for clusters of triangles (right).}}
\vspace{-10px}
\end{wrapfigure}
In the previous section, we employed blending techniques to smoothly replace the costly contact potential $\mathcal{P}^{t_i\cup t_j}$ with a more computationally efficient closed-form potential $\mathcal{P}_c^{t_i\cup t_j}$. However, since this blending is applied only to individual pairs of triangles, the approach still requires summing over $O(T^2)$ terms. In this section, we fully unlock the potential of blending by hierarchically merging triangles to construct a BSH~\citep{agarwal2004collision,bradshaw2004adaptive} for each rigid body, and then smoothly replace the contact potential of each sphere with a single term. Specifically, we adopt a layered hierarchy, where each sphere tightly encapsulates all the bounding spheres of its two children. Another widely used option is to use a wrapped hierarchy, where each sphere tightly encapsulates the actual geometry. Although wrapped hierarchy achieves tighter bound, the layered hierarchy is required in our method to ensure well-behaved properties. Our BSH is defined below:
\begin{define}
\label{def:BSH}
A BSH is a binary tree, where each node contains an index subset of vertices $\mathcal{I}\subseteq\{1,\cdots,N\}$ that is the union of the two subsets of its left and right children, denoted as $\mathcal{I}=\mathcal{I}_l\cup\mathcal{I}_r$. The radius $R_\mathcal{I}$ is the smallest radius encapsulating the bounding spheres of two children. Each leaf node stores a single triangle $t_i$. Further, each node's sphere is centered at $x_\mathcal{I}$ with radius $R_\mathcal{I}$.
\end{define}
Throughout the paper, we use the associated index subset $\mathcal{I}$ to refer to a BSH node. There are many ways to practically construct our BSH, mostly using a greedy algorithm to iteratively merge nodes, and we adopt the technique of~\cite{bradshaw2004adaptive}. Given the BSH, we propose a recursive definition of our contact potential $\mathcal{P}_\text{BSH}^{\mathcal{I}\cup\mathcal{J}}$, one for each node pair $\mathcal{I}$ and $\mathcal{J}$ on two rigid bodies, and use the potential of the root nodes as our final contact potential. Finally, we show that our definition pertains~\prettyref{prop:A},\ref{prop:B},\ref{prop:C}. We start from the base case. For a pair of leaf nodes $t_i\cup t_j$, we use~\prettyref{cor:blending} to define the following pairwise potential:
\begin{align}
\label{eq:leaf-potential}
\begin{cases}
d_1=R_{t_i}+R_{t_j}\text{ and }d_2=(1+\epsilon)d_1\\
\mathcal{P}_{d_1}^{t_i\cup t_j}=\mathcal{P}^{t_i\cup t_j}\text{ and }
\mathcal{P}_{d_2}^{t_i\cup t_j}=\mathcal{P}_c^{t_i\cup t_j}\\
\mathcal{P}_\text{BSH}^{t_i\cup t_j}=\mathcal{P}_{d_1\to d_2}^{t_i\cup t_j}
\end{cases}.
\end{align}
Specifically, we blend the exact potential between the pair of triangles and the closed-form centered potential, when the distance between triangle centers grows by a factor of $\epsilon$. We leave $\epsilon$ as a user-defined margin that controls the exactness of potential evaluation. Next, given an arbitrary internal node with two child nodes being $\mathcal{I}$ and $\mathcal{J}$, we recursively replace the more accurate potential between child nodes with a single potential between parent nodes. Let us suppose $\mathcal{I}\cap\mathcal{J}=\emptyset$, we define a potential of similar form as~\prettyref{eq:centered-potential}:
\begin{align}
\label{eq:centered-cluster-potential}
\mathcal{P}_c^{\mathcal{I}\cap\mathcal{J}}=12\left[1+{1}/{\sqrt{\|x_\mathcal{I}-x_\mathcal{J}\|}}\right]^2,
\end{align}
which is the potential penalizing distance between two sphere centers. We can only use the centered potential when the two spheres are well-separated, i.e. $R_\mathcal{I}+R_\mathcal{J}\leq\|x_\mathcal{I}-x_\mathcal{J}\|$. Otherwise, we have to use the more accurate potential by descending the tree and sum up the pair-wise terms between each pair of child nodes. Specifically, we define the set of child nodes as $C(\mathcal{I})=\{\mathcal{I}_l,\mathcal{I}_r\}$ if $\mathcal{I}$ is an internal node and $C(\mathcal{I})=\{\mathcal{I}\}$ if $\mathcal{I}$ is a leaf node. Finally, we define the following potential between internal node:
\begin{align}
\begin{aligned}
\label{eq:BSH-potential}
&\begin{cases}
d_1=R_\mathcal{I}+R_\mathcal{J}\text{ and }d_2=(1+\epsilon)d_1\\
\mathcal{P}_{d_1}^{\mathcal{I}\cup\mathcal{J}}=\sum_{\mathcal{I}_c\in C(\mathcal{I})}\sum_{\mathcal{J}_c\in C(\mathcal{J})}\mathcal{P}_\text{BSH}^{\mathcal{I}_c\cup\mathcal{J}_c}\text{ and }
\mathcal{P}_{d_2}^{\mathcal{I}\cup\mathcal{J}}=\mathcal{P}_c^{\mathcal{I}\cup\mathcal{J}}\\
\mathcal{P}_\text{BSH}^{\mathcal{I}\cup\mathcal{J}}=
\mathcal{P}_{d_1\to d_2}^{\mathcal{I}\cup\mathcal{J}}
\end{cases}.
\end{aligned}
\end{align}
The main idea behind our formulation is illustrated in~\prettyref{fig:BSH-idea} and we are ready to present our main result, which shows that the so-defined contact potential is well-behaved:
\begin{theorem}
\label{thm:BSH-well-behave}
If $\epsilon>0$ then $\mathcal{P}=\sum_{\mathcal{I}\neq\mathcal{J}}\mathcal{P}_\text{BSH}^{\mathcal{I}\cup\mathcal{J}}$ pertains~\prettyref{prop:A},\ref{prop:B},\ref{prop:C}, where the summation is taken over the root nodes of different rigid bodies.
\end{theorem}
This result lays the foundation for our efficient-to-evaluate and well-behaved contact model. Although analyzing the cost of evaluating $\mathcal{P}$ could be rather difficult for general cases, we follow the idea of fast multiple expansion~\citep{greengard1987fast} and analyze the cost of evaluating the contact potential for a uniform grid, where we show in~\prettyref{appen:complexity} that the cost is $O(T)$. Finally, we notice that a contact model should account for frictional contact forces. In~\prettyref{appen:fri}, we show that the frictional damping potential proposed in~\cite{ye2024sdrs} can be slightly extended to ours.
\section{\label{sec:evaluation}Evaluation}
We evaluated our method in a row of five contact-rich manipulation and locomotion tasks: Billiards, Push, Sort, Ant-Push and Gather. We optimize the sequence of control signals using gradient descent at a fixed learning rate to minimize user-defined loss functions. More experiments and details are deferred to~\prettyref{appen:experiment}. For fairness, we compare our contact model with the standard IPC model used in~\cite{li2020incremental,huang2024differentiable}, and SDRS contact model proposed by~\cite{ye2024sdrs}, which only
doesn't satisfy \prettyref{prop:C}b. We also compare with~\cite{suh2022bundled} that uses first-order bundled gradient.

\begin{figure*}[ht]
\centering
\scalebox{.63}{
\setlength{\tabcolsep}{1px}
\begin{tabular}{cccc}
\includegraphics[height=.24\linewidth,frame,
trim=1cm 5cm 4cm 6cm,clip]{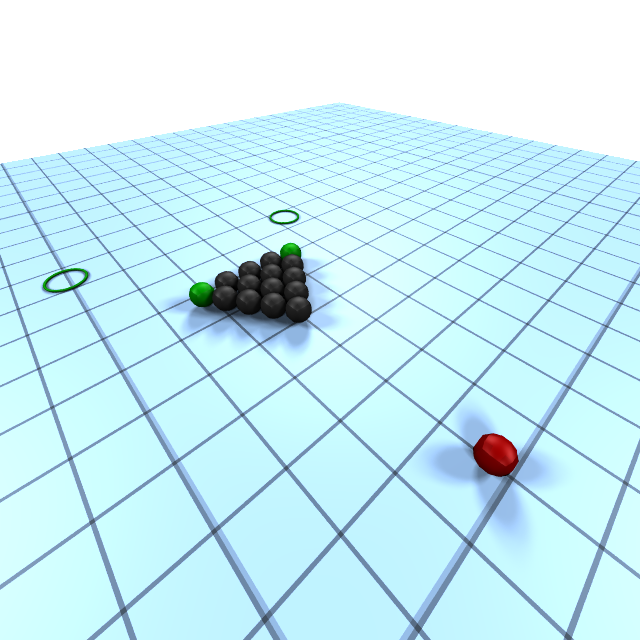}
\put(-50,5){(a)}&
\includegraphics[height=.24\linewidth,frame,
trim=1cm 5cm 4cm 6cm,clip]{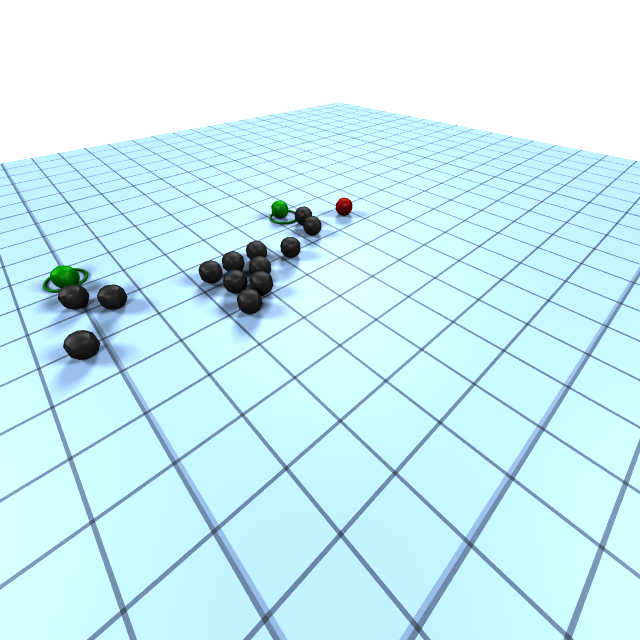}
\put(-50,5){(b)}&
\includegraphics[height=.24\linewidth]{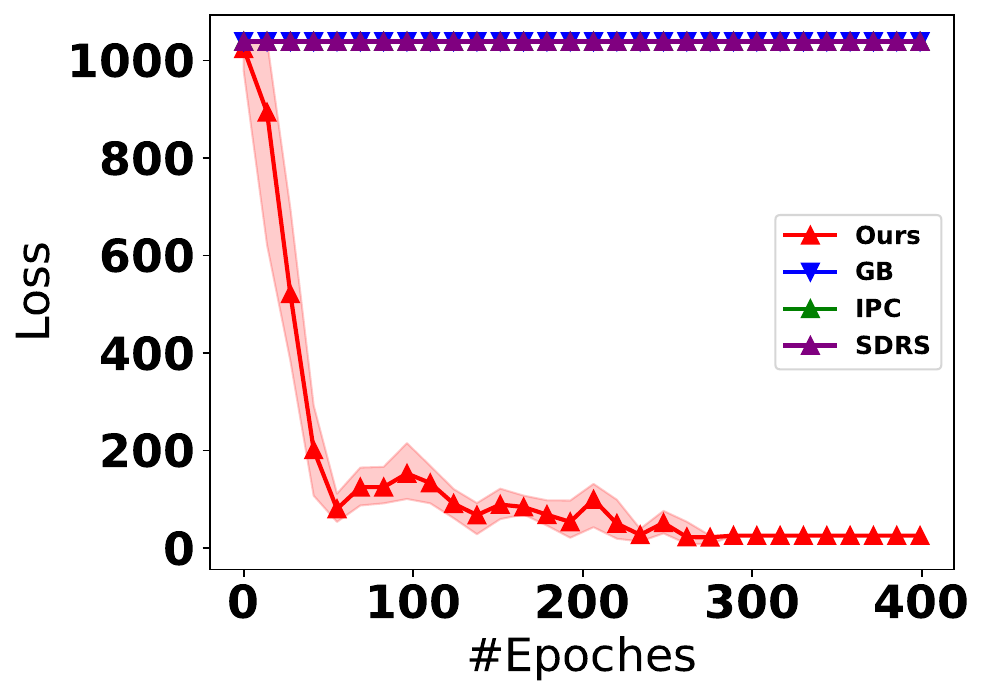}
\put(-20,75){(c)}&
\includegraphics[height=.24\linewidth]{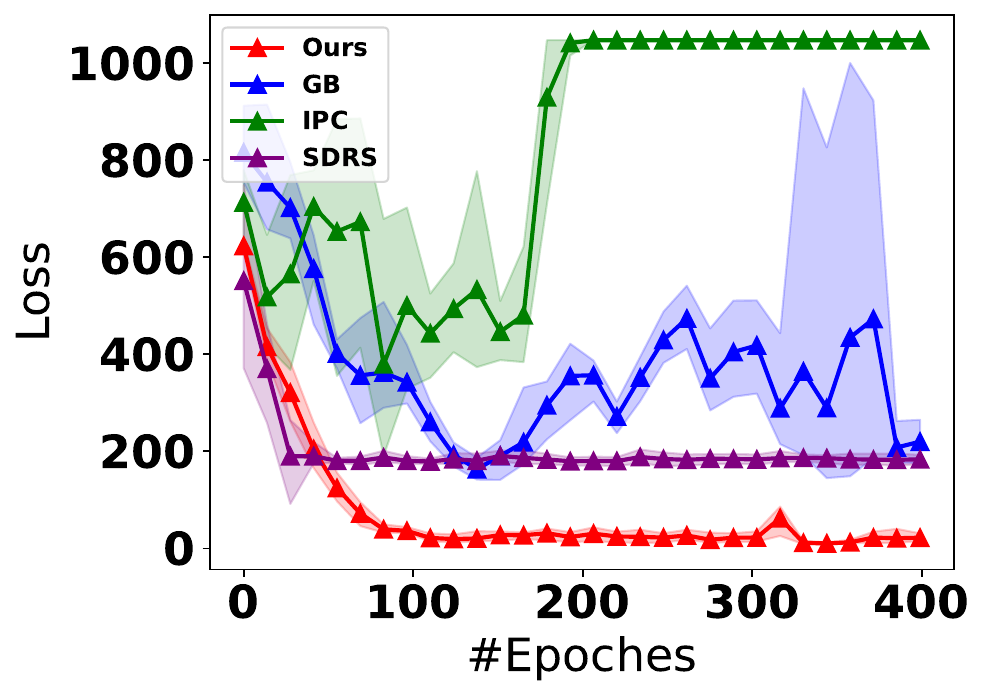}
\put(-25,75){(d)}
\end{tabular}}
\caption{\label{fig:billiards}\footnotesize{We show the initial (a) and final (b) frame of the billiards task, the convergence history uses trivial initialization (c) and random sampling initialization (d).}}
\end{figure*}
\begin{wrapfigure}{r}{0.33\linewidth}
\centering
\vspace{-10px}
\setlength{\tabcolsep}{5px}
\begin{tabular}{cc} 
\includegraphics[height=.4\linewidth]{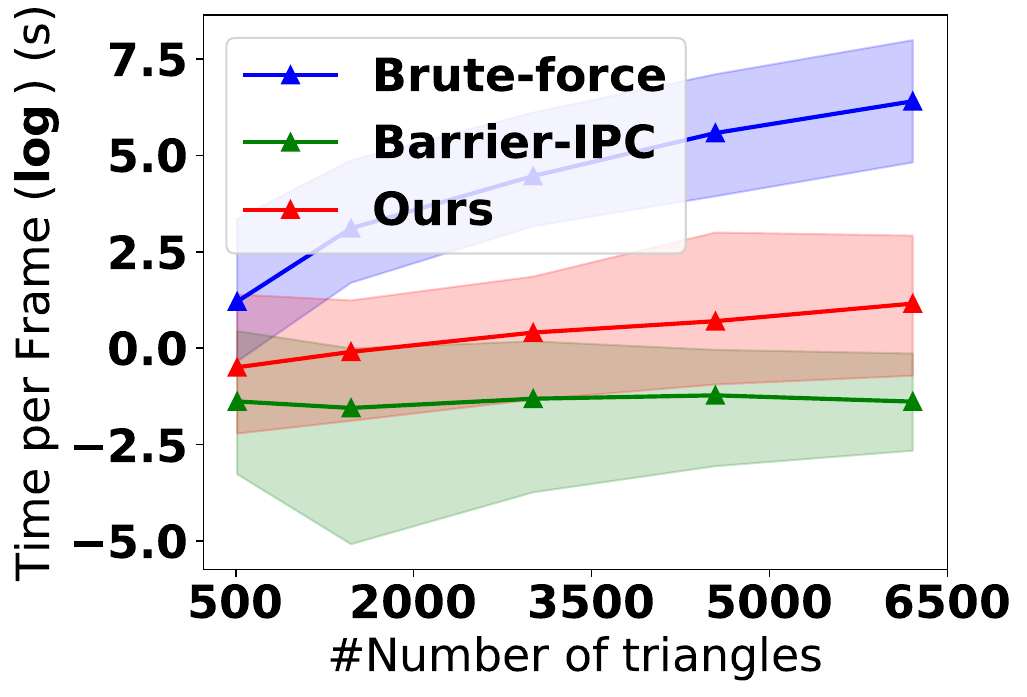} & 
\includegraphics[height=.4\linewidth]{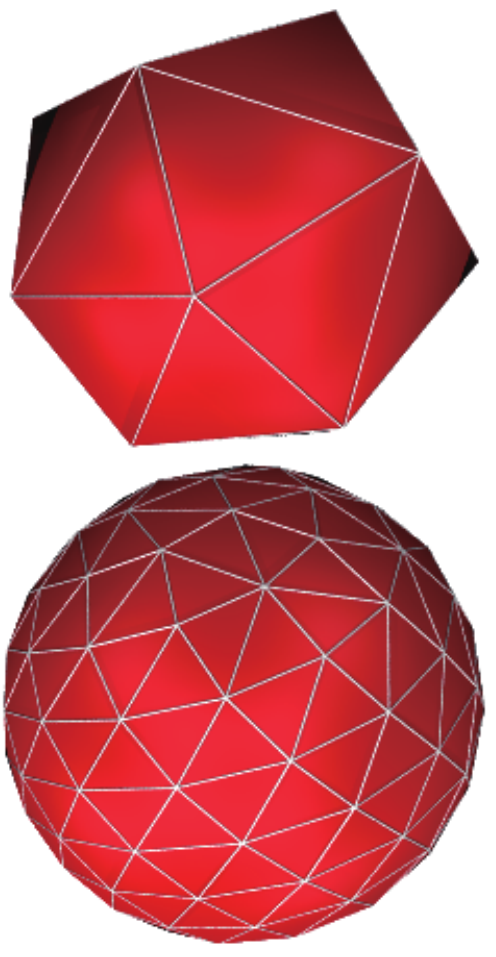} 
\end{tabular}
\vspace{-5px}
\caption{\label{fig:bunny}\footnotesize{The cost of computing contact potential (left) under different ball mesh resolutions (right).}}
\vspace{-10px}
\end{wrapfigure}
\textbf{Billiards:} In this benchmark, we have 16 balls on the ground with one target red ball whose initial horizontal position and velocity can be controlled. The goal of control is for the two green balls to reach the target positions (green circle), where the loss function is the squared distance between the green balls and the center of green circles. We  experiment with two different methods for setting initial solutions. Our first method uses trivial initialization where certain rigid objects in a scenario are far apart, for which the gradient might vanish except our method. Our second method uses random sampling of control signals to find an initial solution for which gradient information does not vanish. The convergence history of various contact models and initialization strategies is summarized in~\prettyref{fig:billiards}. We optimize a trajectory with $100$ timesteps at a timestep size of $0.04$. Expect our method, other methods cannot make any progress without sampling due to gradient vanishing, which can be fixed via sampling, while our method achieves faster convergence, with or without sampling. We further pick random frames in this scenario and compare the cost of computing the contact potential using our method and IPC. Clearly, our method is slower than IPC due to nested optimization and hierarchical blending. But our method accelerated by BSH is much faster than the brute-force version computing all pairwise potentials between triangles.
\begin{figure*}[ht]
\centering
\scalebox{.53}{
\setlength{\tabcolsep}{1px}
\begin{tabular}{cccc}
\includegraphics[height=.24\linewidth,frame,
trim=2cm 6cm 2cm 8cm,clip]{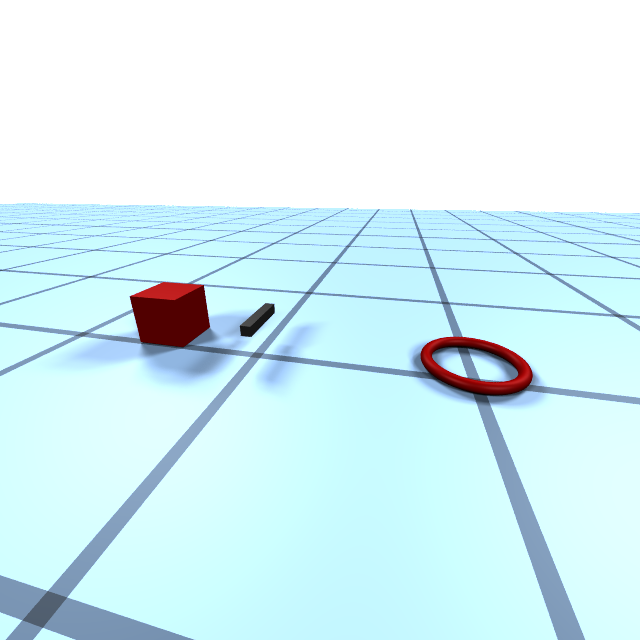}
\put(-50,5){(a)}&
\includegraphics[height=.24\linewidth,frame,
trim=2cm 6cm 2cm 8cm,clip]{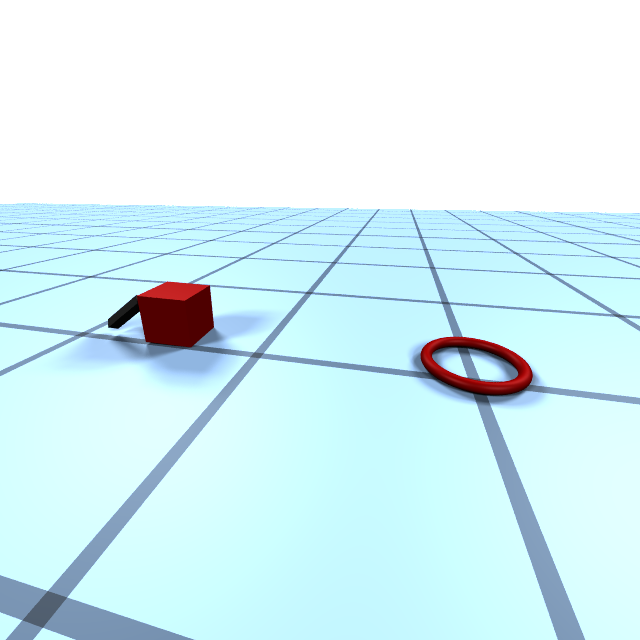}
\put(-50,5){(b)}&
\includegraphics[height=.24\linewidth,frame,
trim=2cm 6cm 2cm 8cm,clip]{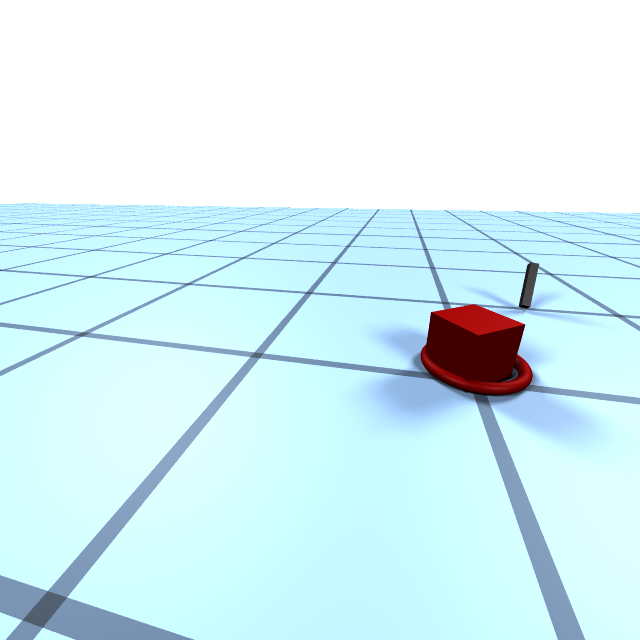}
\put(-50,5){(c)}&
\includegraphics[height=.24\linewidth]{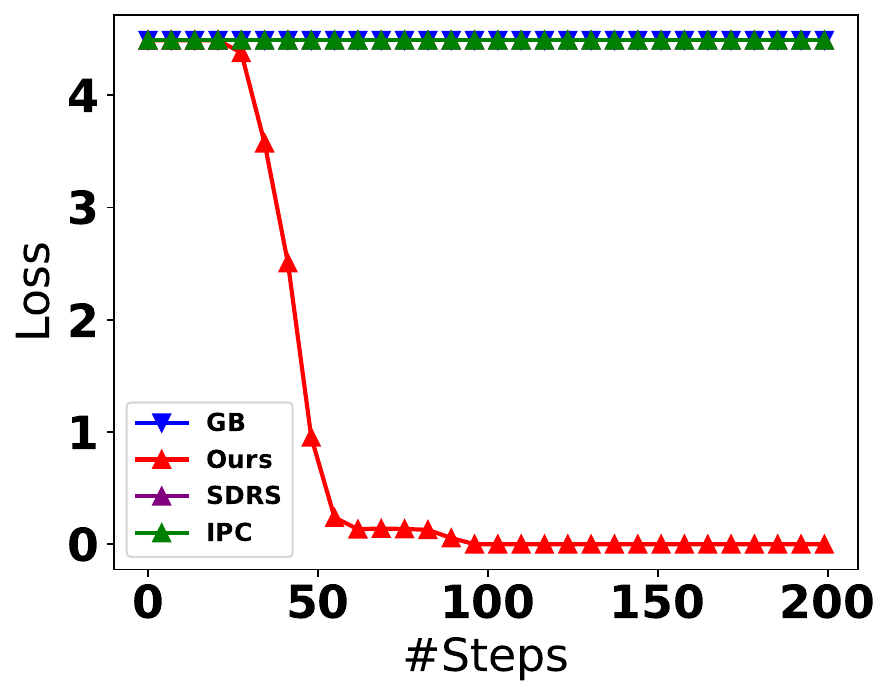}
\put(-50,75){(d)}
\end{tabular}}
\caption{\label{fig:push}\footnotesize{We show the initial (a), middle (b), final (c) frame of the push task, and the convergence history (d) of our method, IPC-based contact model, and gradient bundle method (GB).}}
\end{figure*}
\textbf{Push:} In this benchmark, we optimize the position and orientation of a rod to push the red box to reach a target red circle, where the loss is the squared distance between the box and the red circle. For this benchmark, we use receding-horizon control to generate a trajectory of $200$ frames with a horizon of only $48$ frames, i.e., we iteratively optimize a $48$-frame sub-trajectory and only apply the first action. The rod is initially located in front of the box, and our contact model guides the rod to first move around to the back of the box, and then push it multiple times to reach the target area as illustrated in~\prettyref{fig:push}. This result demonstrates the ability of our contact model to discover multi-stage, contact-rich motions from a trivial initial guess. In comparison, all other methods make no progress, even with random sampling to find an initial solution that gradient does not vanish.
\begin{figure*}[ht]
\centering
\scalebox{.53}{
\setlength{\tabcolsep}{1px}
\begin{tabular}{cccc}
\includegraphics[height=.24\linewidth,frame,
trim=8cm 8cm 6cm 10cm,clip]{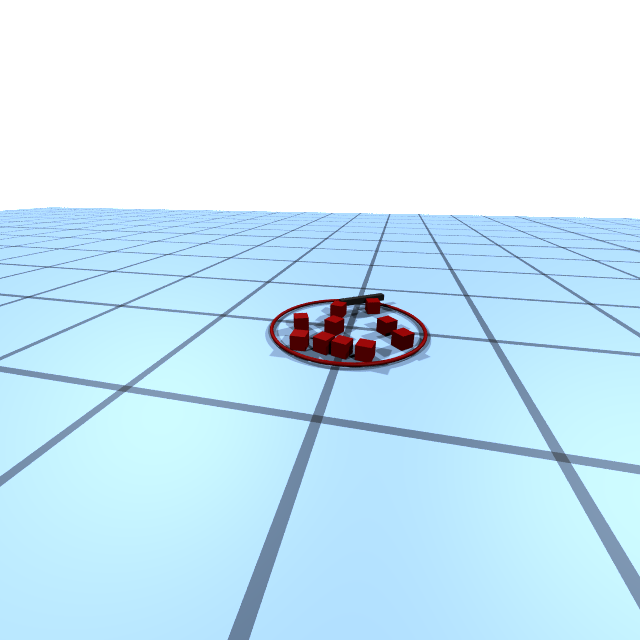}
\put(-50,5){(a)}&
\includegraphics[height=.24\linewidth,frame,
trim=0cm 6cm 0cm 8cm,clip]{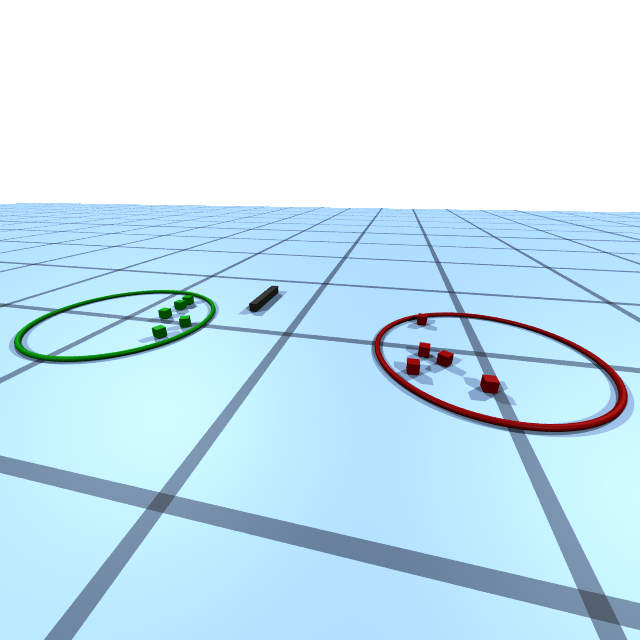}
\put(-70,5){(b)}&
\includegraphics[height=.24\linewidth]{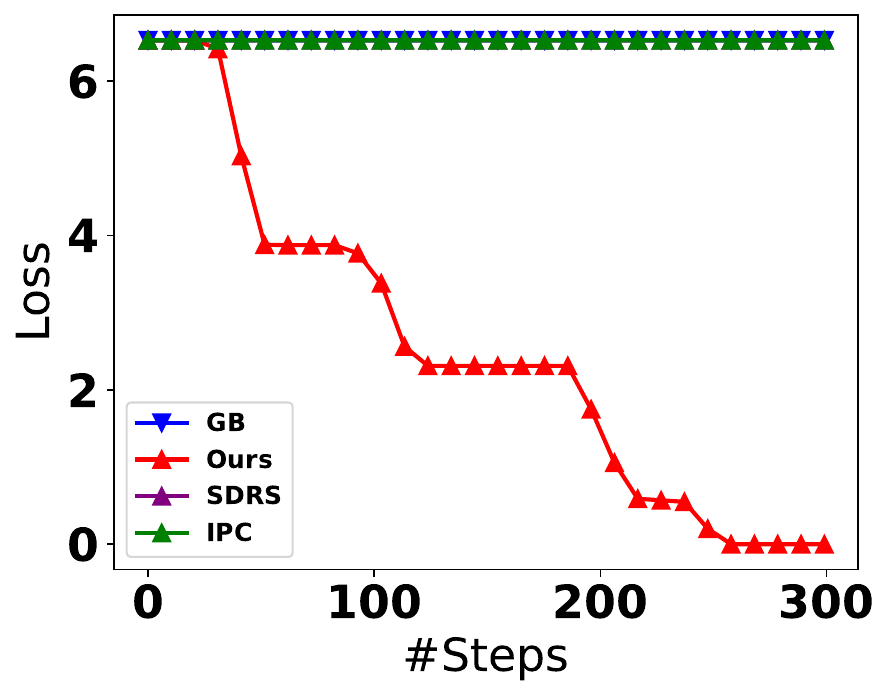}
\put(-50,55){(c)}&
\includegraphics[height=.24\linewidth]{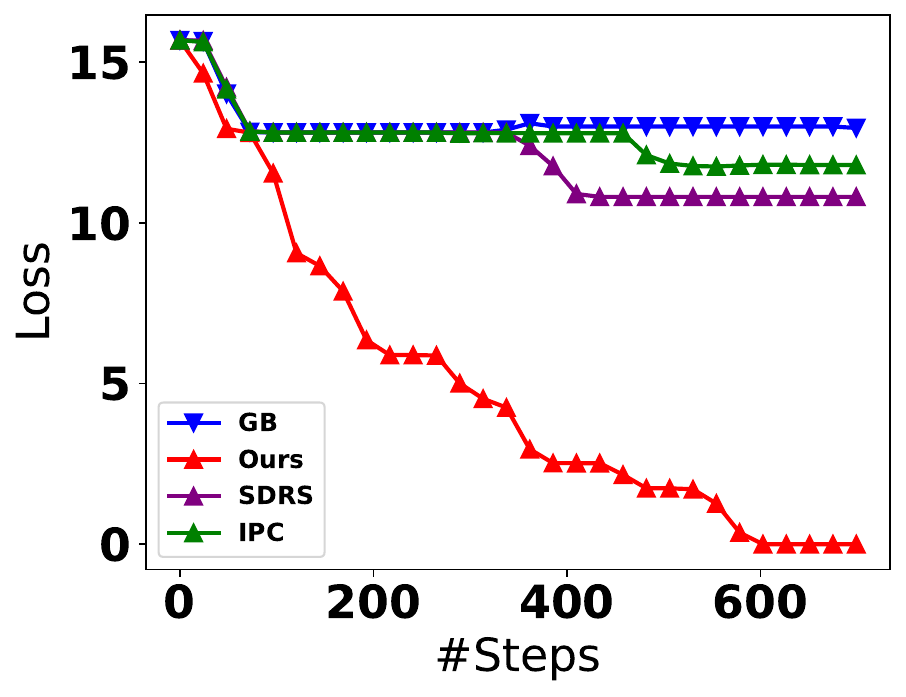}
\put(-50,55){(d)}
\end{tabular}}
\caption{\label{fig:gather-sort}\footnotesize{The final frame of the gather (a) and sort (b) task, with convergence history in (c) and (d).}}
\end{figure*}
\textbf{Gather \& Sort:} In these two benchmarks, we further increase the complexity to have the rod manipulate multiple objects on the table. In the gather task, $10$ cubes are disseminated and to be pushed together into the target region. In the sort task, $10$ cubes are randomly labeled and to be pushed into separate regions according to their labels. The loss is the sum of squared distance between boxes and targets. The frames and convergence history are illustrated in~\prettyref{fig:gather-sort}. Again, other methods make little to no progress, while our method successfully accomplishes the task, relying solely on the gradient information.
\begin{figure*}[ht]
\centering
\scalebox{.53}{
\setlength{\tabcolsep}{1px}
\begin{tabular}{ccc}
\includegraphics[height=.24\linewidth,frame,
trim=1cm 6cm 2cm 8cm,clip]{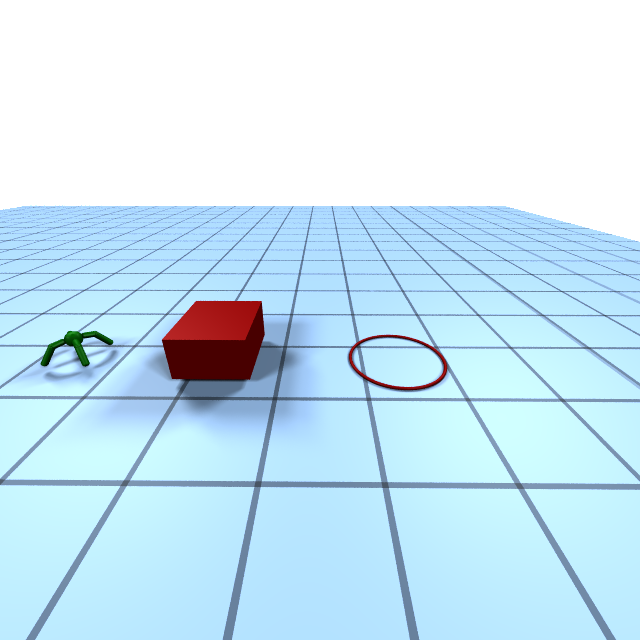}
\put(-50,5){(a)}&
\includegraphics[height=.24\linewidth,frame,
trim=1cm 6cm 2cm 8cm,clip]{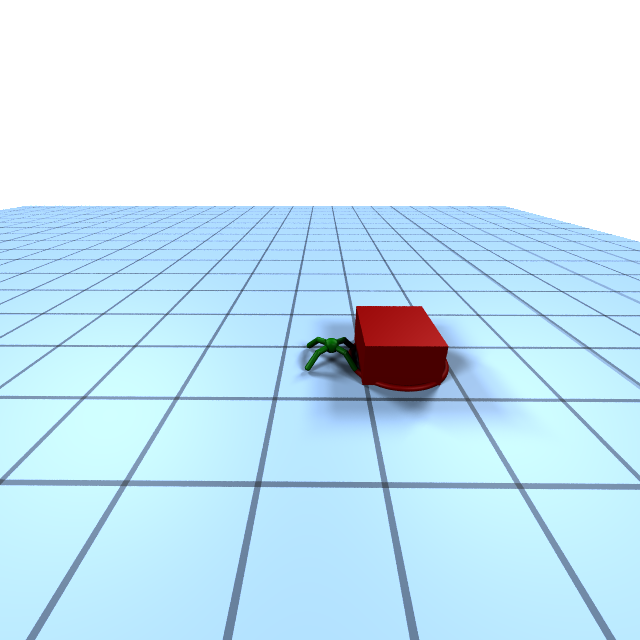}
\put(-50,5){(b)}&
\includegraphics[height=.24\linewidth]{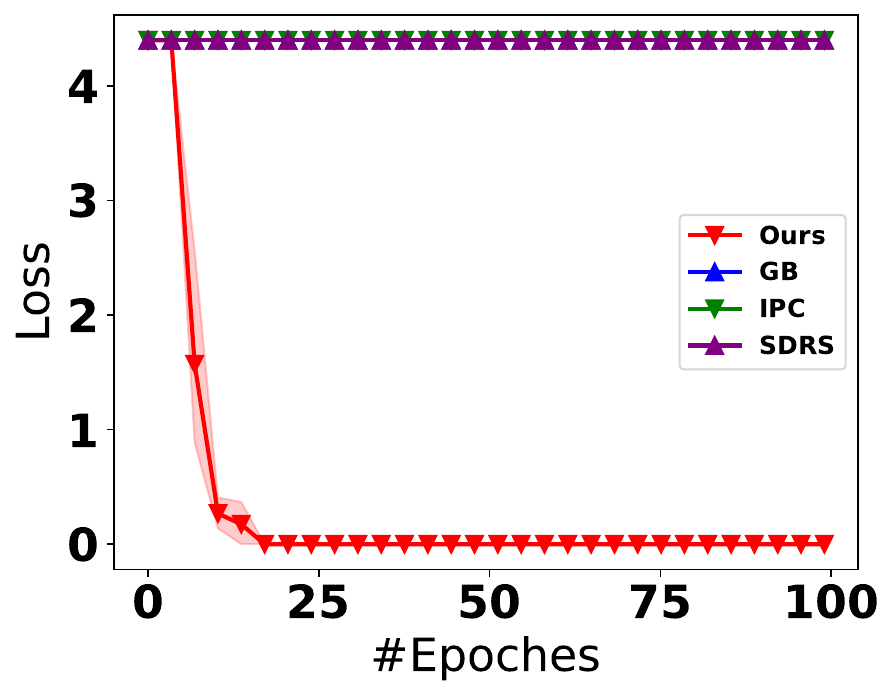}
\put(-50,75){(c)}
\end{tabular}}
\caption{\label{fig:spider}\footnotesize{The initial (a), final (b) frame of the Ant-Push task, and the convergence history (c).}}
\end{figure*}
\textbf{Ant-Push:} In this benchmark, we show that our method can work with articulated robot models. We optimize for a 9-link, 16-DOF ant robot to push a box to the target position. Once again, our method significantly outperforms other methods, as illustrated int \prettyref{fig:spider}.
\section{Conclusion}
We present a detailed analysis of the qualifications for a contact model to be well-behaved, which strictly prevents collisions, supports differentiable simulations, induce non-prehensile forces, and avoids vanishing gradients. By hierarchically evaluating the contact potentials assisted by a BSH, we further present a well-behaved contact model that is also efficient to evaluate. By analysis on the special case of a uniform grid, we show that the complexity of evaluating our contact potential is linear. Through evaluations on various motion planning and control tasks, we highlight that our model can guide a gradient-based optimizer to search for complex motion plans and locomotion gaits that are impossible for previous contact models. 
Our method is not without its problems. First, we can only handle rigid bodies, and we cannot deal with more general deformable objects for soft robot locomotion or soft object manipulation. This is because our bounding spheres might not bound the actual triangles if deformation happens, which could potentially violate~\prettyref{prop:C}. Second, our contact potential involves a recursive definition and requires a nested optimization between pairs of triangles, which incurs considerable overhead to a conventional rigid body simulator. 

\bibliography{template}

\begin{thebibliography}{42}
\providecommand{\natexlab}[1]{#1}
\providecommand{\url}[1]{\texttt{#1}}
\expandafter\ifx\csname urlstyle\endcsname\relax
  \providecommand{\doi}[1]{doi: #1}\else
  \providecommand{\doi}{doi: \begingroup \urlstyle{rm}\Url}\fi

\bibitem[Agarwal et~al.(2004)Agarwal, Guibas, Nguyen, Russel, and Zhang]{agarwal2004collision}
Pankaj Agarwal, Leonidas Guibas, An~Nguyen, Daniel Russel, and Li~Zhang.
\newblock Collision detection for deforming necklaces.
\newblock \emph{Computational Geometry}, 28\penalty0 (2-3):\penalty0 137--163, 2004.

\bibitem[Amos et~al.(2018)Amos, Jimenez, Sacks, Boots, and Kolter]{amos2018differentiable}
Brandon Amos, Ivan Jimenez, Jacob Sacks, Byron Boots, and J~Zico Kolter.
\newblock Differentiable mpc for end-to-end planning and control.
\newblock \emph{Advances in neural information processing systems}, 31, 2018.

\bibitem[Antonova et~al.(2023)Antonova, Yang, Jatavallabhula, and Bohg]{antonova2023rethinking}
Rika Antonova, Jingyun Yang, Krishna~Murthy Jatavallabhula, and Jeannette Bohg.
\newblock Rethinking optimization with differentiable simulation from a global perspective.
\newblock In \emph{Conference on Robot Learning}, pp.\  276--286. PMLR, 2023.

\bibitem[Barnes \& Hut(1986)Barnes and Hut]{barnes1986hierarchical}
Josh Barnes and Piet Hut.
\newblock A hierarchical o (n log n) force-calculation algorithm.
\newblock \emph{nature}, 324\penalty0 (6096):\penalty0 446--449, 1986.

\bibitem[Bradshaw \& O'Sullivan(2004)Bradshaw and O'Sullivan]{bradshaw2004adaptive}
Gareth Bradshaw and Carol O'Sullivan.
\newblock Adaptive medial-axis approximation for sphere-tree construction.
\newblock \emph{ACM Transactions on Graphics (TOG)}, 23\penalty0 (1):\penalty0 1--26, 2004.

\bibitem[Brochu et~al.(2012)Brochu, Edwards, and Bridson]{brochu2012efficient}
Tyson Brochu, Essex Edwards, and Robert Bridson.
\newblock Efficient geometrically exact continuous collision detection.
\newblock \emph{ACM Transactions on Graphics (TOG)}, 31\penalty0 (4):\penalty0 1--7, 2012.

\bibitem[de~Avila Belbute-Peres et~al.(2018)de~Avila Belbute-Peres, Smith, Allen, Tenenbaum, and Kolter]{de2018end}
Filipe de~Avila Belbute-Peres, Kevin Smith, Kelsey Allen, Josh Tenenbaum, and J~Zico Kolter.
\newblock End-to-end differentiable physics for learning and control.
\newblock \emph{Advances in neural information processing systems}, 31, 2018.

\bibitem[Dontchev \& Rockafellar(2009)Dontchev and Rockafellar]{dontchev2009implicit}
Asen~L Dontchev and R~Tyrrell Rockafellar.
\newblock \emph{Implicit functions and solution mappings}, volume 543.
\newblock Springer, 2009.

\bibitem[Du et~al.(2021)Du, Wu, Ma, Wah, Spielberg, Rus, and Matusik]{Du2021diffpd}
Tao Du, Kui Wu, Pingchuan Ma, Sebastien Wah, Andrew Spielberg, Daniela Rus, and Wojciech Matusik.
\newblock Diffpd: Differentiable projective dynamics.
\newblock \emph{ACM Trans. Graph.}, 41\penalty0 (2), November 2021.
\newblock ISSN 0730-0301.

\bibitem[Fisher \& Lin(2001)Fisher and Lin]{fisher2001deformed}
Susan Fisher and Ming~C Lin.
\newblock Deformed distance fields for simulation of non-penetrating flexible bodies.
\newblock In \emph{Computer Animation and Simulation 2001: Proceedings of the Eurographics Workshop in Manchester, UK, September 2--3, 2001}, pp.\  99--111. Springer, 2001.

\bibitem[Gast et~al.(2015)Gast, Schroeder, Stomakhin, Jiang, and Teran]{gast2015optimization}
Theodore~F Gast, Craig Schroeder, Alexey Stomakhin, Chenfanfu Jiang, and Joseph~M Teran.
\newblock Optimization integrator for large time steps.
\newblock \emph{IEEE transactions on visualization and computer graphics}, 21\penalty0 (10):\penalty0 1103--1115, 2015.

\bibitem[Greengard \& Rokhlin(1987)Greengard and Rokhlin]{greengard1987fast}
Leslie Greengard and Vladimir Rokhlin.
\newblock A fast algorithm for particle simulations.
\newblock \emph{Journal of computational physics}, 73\penalty0 (2):\penalty0 325--348, 1987.

\bibitem[Guendelman et~al.(2003)Guendelman, Bridson, and Fedkiw]{guendelman2003nonconvex}
Eran Guendelman, Robert Bridson, and Ronald Fedkiw.
\newblock Nonconvex rigid bodies with stacking.
\newblock \emph{ACM transactions on graphics (TOG)}, 22\penalty0 (3):\penalty0 871--878, 2003.

\bibitem[Harmon et~al.(2009)Harmon, Vouga, Smith, Tamstorf, and Grinspun]{harmon2009asynchronous}
David Harmon, Etienne Vouga, Breannan Smith, Rasmus Tamstorf, and Eitan Grinspun.
\newblock Asynchronous contact mechanics.
\newblock In \emph{ACM SIGGRAPH 2009 papers}, pp.\  1--12. 2009.

\bibitem[Heiden et~al.(2021)Heiden, Millard, Coumans, Sheng, and Sukhatme]{heiden2021neuralsim}
Eric Heiden, David Millard, Erwin Coumans, Yizhou Sheng, and Gaurav~S Sukhatme.
\newblock Neuralsim: Augmenting differentiable simulators with neural networks.
\newblock In \emph{2021 IEEE International Conference on Robotics and Automation (ICRA)}, pp.\  9474--9481. IEEE, 2021.

\bibitem[Hu et~al.(2019)Hu, Liu, Spielberg, Tenenbaum, Freeman, Wu, Rus, and Matusik]{hu2019chainqueen}
Yuanming Hu, Jiancheng Liu, Andrew Spielberg, Joshua~B Tenenbaum, William~T Freeman, Jiajun Wu, Daniela Rus, and Wojciech Matusik.
\newblock Chainqueen: A real-time differentiable physical simulator for soft robotics.
\newblock In \emph{2019 International conference on robotics and automation (ICRA)}, pp.\  6265--6271. IEEE, 2019.

\bibitem[Huang et~al.(2024)Huang, Tozoni, Gjoka, Ferguson, Schneider, Panozzo, and Zorin]{huang2024differentiable}
Zizhou Huang, Davi~Colli Tozoni, Arvi Gjoka, Zachary Ferguson, Teseo Schneider, Daniele Panozzo, and Denis Zorin.
\newblock Differentiable solver for time-dependent deformation problems with contact.
\newblock \emph{ACM Transactions on Graphics}, 43\penalty0 (3):\penalty0 1--30, 2024.

\bibitem[Kingma \& Ba(2014)Kingma and Ba]{kingma2014adam}
Diederik~P Kingma and Jimmy Ba.
\newblock Adam: A method for stochastic optimization.
\newblock \emph{arXiv preprint arXiv:1412.6980}, 2014.

\bibitem[Le~Cleac'h et~al.(2023)Le~Cleac'h, Schwager, Manchester, Sindhwani, Florence, and Singh]{10105986}
Simon Le~Cleac'h, Mac Schwager, Zachary Manchester, Vikas Sindhwani, Pete Florence, and Sumeet Singh.
\newblock Single-level differentiable contact simulation.
\newblock \emph{IEEE Robotics and Automation Letters}, 8\penalty0 (7):\penalty0 4012--4019, 2023.
\newblock \doi{10.1109/LRA.2023.3268824}.

\bibitem[Le~Lidec et~al.(2021)Le~Lidec, Kalevatykh, Laptev, Schmid, and Carpentier]{le2021differentiable}
Quentin Le~Lidec, Igor Kalevatykh, Ivan Laptev, Cordelia Schmid, and Justin Carpentier.
\newblock Differentiable simulation for physical system identification.
\newblock \emph{IEEE Robotics and Automation Letters}, 6\penalty0 (2):\penalty0 3413--3420, 2021.

\bibitem[Levine \& Koltun(2013)Levine and Koltun]{levine2013guided}
Sergey Levine and Vladlen Koltun.
\newblock Guided policy search.
\newblock In \emph{International conference on machine learning}, pp.\  1--9. PMLR, 2013.

\bibitem[Li et~al.(2020)Li, Ferguson, Schneider, Langlois, Zorin, Panozzo, Jiang, and Kaufman]{li2020incremental}
Minchen Li, Zachary Ferguson, Teseo Schneider, Timothy~R Langlois, Denis Zorin, Daniele Panozzo, Chenfanfu Jiang, and Danny~M Kaufman.
\newblock Incremental potential contact: intersection-and inversion-free, large-deformation dynamics.
\newblock \emph{ACM Trans. Graph.}, 39\penalty0 (4):\penalty0 49, 2020.

\bibitem[Li et~al.(2022{\natexlab{a}})Li, Huang, Du, Su, Tenenbaum, and Gan]{li2022contact}
Sizhe Li, Zhiao Huang, Tao Du, Hao Su, Joshua Tenenbaum, and Chuang Gan.
\newblock {C}ontact {P}oints {D}iscovery for {S}oft-{B}ody {M}anipulations with {D}ifferentiable {P}hysics.
\newblock In \emph{International Conference on Learning Representations (ICLR)}, 2022{\natexlab{a}}.

\bibitem[Li et~al.(2022{\natexlab{b}})Li, Du, Wu, Xu, and Matusik]{Li2022DiffCloth}
Yifei Li, Tao Du, Kui Wu, Jie Xu, and Wojciech Matusik.
\newblock Diffcloth: Differentiable cloth simulation with dry frictional contact.
\newblock \emph{ACM Trans. Graph.}, 42\penalty0 (1), October 2022{\natexlab{b}}.
\newblock ISSN 0730-0301.

\bibitem[Li et~al.(2023)Li, Xu, Ye, Ren, and Liu]{li2023difffr}
Zhehao Li, Qingyu Xu, Xiaohan Ye, Bo~Ren, and Ligang Liu.
\newblock Difffr: Differentiable sph-based fluid-rigid coupling for rigid body control.
\newblock \emph{ACM Transactions on Graphics (TOG)}, 42\penalty0 (6):\penalty0 1--17, 2023.

\bibitem[Liang et~al.(2024)Liang, Gao, Wu, and Pan]{liang2024second}
Chen Liang, Xifeng Gao, Kui Wu, and Zherong Pan.
\newblock Second-order convergent collision-constrained optimization-based planner.
\newblock \emph{IEEE Robotics and Automation Letters}, 2024.

\bibitem[Ma et~al.(2022)Ma, Du, Tenenbaum, Matusik, and Gan]{ma2022risp}
Pingchuan Ma, Tao Du, Joshua~B. Tenenbaum, Wojciech Matusik, and Chuang Gan.
\newblock {RISP}: Rendering-invariant state predictor with differentiable simulation and rendering for cross-domain parameter estimation.
\newblock In \emph{International Conference on Learning Representations}, 2022.
\newblock URL \url{https://openreview.net/forum?id=uSE03demja}.

\bibitem[Marsden \& West(2001)Marsden and West]{marsden2001discrete}
Jerrold~E Marsden and Matthew West.
\newblock Discrete mechanics and variational integrators.
\newblock \emph{Acta numerica}, 10:\penalty0 357--514, 2001.

\bibitem[Mordatch et~al.(2012)Mordatch, Todorov, and Popovi{\'c}]{mordatch2012discovery}
Igor Mordatch, Emanuel Todorov, and Zoran Popovi{\'c}.
\newblock Discovery of complex behaviors through contact-invariant optimization.
\newblock \emph{ACM Transactions on Graphics (ToG)}, 31\penalty0 (4):\penalty0 1--8, 2012.

\bibitem[Newbury et~al.(2024)Newbury, Collins, He, Pan, Posner, Howard, and Cosgun]{newbury2024review}
Rhys Newbury, Jack Collins, Kerry He, Jiahe Pan, Ingmar Posner, David Howard, and Akansel Cosgun.
\newblock A review of differentiable simulators.
\newblock \emph{IEEE Access}, 2024.

\bibitem[Pan \& Manocha(2018)Pan and Manocha]{pan2018active}
Zherong Pan and Dinesh Manocha.
\newblock Active animations of reduced deformable models with environment interactions.
\newblock \emph{ACM Transactions on Graphics (TOG)}, 37\penalty0 (3):\penalty0 1--17, 2018.

\bibitem[Pang et~al.(2023)Pang, Suh, Yang, and Tedrake]{pang2023global}
Tao Pang, HJ~Terry Suh, Lujie Yang, and Russ Tedrake.
\newblock Global planning for contact-rich manipulation via local smoothing of quasi-dynamic contact models.
\newblock \emph{IEEE Transactions on robotics}, 2023.

\bibitem[Stuyck \& Chen(2023)Stuyck and Chen]{stuyck2023diffxpbd}
Tuur Stuyck and Hsiao-yu Chen.
\newblock Diffxpbd: Differentiable position-based simulation of compliant constraint dynamics.
\newblock \emph{Proceedings of the ACM on Computer Graphics and Interactive Techniques}, 6\penalty0 (3):\penalty0 1--14, 2023.

\bibitem[Suh et~al.(2022{\natexlab{a}})Suh, Simchowitz, Zhang, and Tedrake]{suh2022differentiable}
Hyung~Ju Suh, Max Simchowitz, Kaiqing Zhang, and Russ Tedrake.
\newblock Do differentiable simulators give better policy gradients?
\newblock In \emph{International Conference on Machine Learning}, pp.\  20668--20696. PMLR, 2022{\natexlab{a}}.

\bibitem[Suh et~al.(2022{\natexlab{b}})Suh, Pang, and Tedrake]{suh2022bundled}
Hyung Ju~Terry Suh, Tao Pang, and Russ Tedrake.
\newblock Bundled gradients through contact via randomized smoothing.
\newblock \emph{IEEE Robotics and Automation Letters}, 7\penalty0 (2):\penalty0 4000--4007, 2022{\natexlab{b}}.

\bibitem[Tassa et~al.(2012)Tassa, Erez, and Todorov]{6386025}
Yuval Tassa, Tom Erez, and Emanuel Todorov.
\newblock Synthesis and stabilization of complex behaviors through online trajectory optimization.
\newblock In \emph{2012 IEEE/RSJ International Conference on Intelligent Robots and Systems}, pp.\  4906--4913, 2012.
\newblock \doi{10.1109/IROS.2012.6386025}.

\bibitem[Todorov(2011)]{5979814}
Emanuel Todorov.
\newblock A convex, smooth and invertible contact model for trajectory optimization.
\newblock In \emph{2011 IEEE International Conference on Robotics and Automation}, pp.\  1071--1076, 2011.
\newblock \doi{10.1109/ICRA.2011.5979814}.

\bibitem[Toussaint et~al.(2019)Toussaint, Allen, Smith, and Tenenbaum]{ijcai2019p869}
Marc Toussaint, Kelsey~R. Allen, Kevin~A. Smith, and Joshua~B. Tenenbaum.
\newblock Differentiable physics and stable modes for tool-use and manipulation planning - extended abtract.
\newblock In \emph{Proceedings of the Twenty-Eighth International Joint Conference on Artificial Intelligence, {IJCAI-19}}, pp.\  6231--6235. International Joint Conferences on Artificial Intelligence Organization, 7 2019.
\newblock \doi{10.24963/ijcai.2019/869}.
\newblock URL \url{https://doi.org/10.24963/ijcai.2019/869}.

\bibitem[Werling et~al.(2021)Werling, Omens, Lee, Exarchos, and Liu]{Werling-RSS-21}
Keenon Werling, Dalton Omens, Jeongseok Lee, Ioannis Exarchos, and C.~Karen Liu.
\newblock {Fast and Feature-Complete Differentiable Physics Engine for Articulated Rigid Bodies with Contact Constraints}.
\newblock In \emph{Proceedings of Robotics: Science and Systems}, Virtual, July 2021.
\newblock \doi{10.15607/RSS.2021.XVII.034}.

\bibitem[Xu et~al.(2021)Xu, Chen, Zlokapa, Foshey, Matusik, Sueda, and Agrawal]{Xu-RSS-21}
Jie Xu, Tao Chen, Lara Zlokapa, Michael Foshey, Wojciech Matusik, Shinjiro Sueda, and Pulkit Agrawal.
\newblock {An End-to-End Differentiable Framework for Contact-Aware Robot Design}.
\newblock In \emph{Proceedings of Robotics: Science and Systems}, Virtual, July 2021.
\newblock \doi{10.15607/RSS.2021.XVII.008}.

\bibitem[Xu et~al.(2022)Xu, Macklin, Makoviychuk, Narang, Garg, Ramos, and Matusik]{xu2022accelerated}
Jie Xu, Miles Macklin, Viktor Makoviychuk, Yashraj Narang, Animesh Garg, Fabio Ramos, and Wojciech Matusik.
\newblock Accelerated policy learning with parallel differentiable simulation.
\newblock In \emph{International Conference on Learning Representations}, 2022.
\newblock URL \url{https://openreview.net/forum?id=ZSKRQMvttc}.

\bibitem[Ye et~al.(2024)Ye, Gao, Wu, Pan, and Komura]{ye2024sdrs}
Xiaohan Ye, Xifeng Gao, Kui Wu, Zherong Pan, and Taku Komura.
\newblock Sdrs: Shape-differentiable robot simulator.
\newblock \emph{arXiv preprint arXiv:2412.19127}, 2024.

\end{thebibliography}
\bibliographystyle{iclr2026_conference}
\newpage
\appendix
\section{Appendix}
\subsection{\label{appen:proof}Additional Proofs}
\begin{proof}[proof of~\prettyref{lem:slow}]
\TE{\prettyref{prop:A}} If $x\in\CFree$ then each pair of $t_i,t_j$ on different rigid bodies is disjoint as the convex hulls are closed sets. Therefore, by the separating plane theorem, there exists a separating plane $p_{ij}$ and some positive $\epsilon_{ij}>0$ such that:
\begin{align*}
\|n_{ij}\|=1\land
\langle x_{i(k)\in t_i},n_{ij}\rangle+d_{ij}\geq\epsilon_{ij}/2\land
\langle x_{j(k)\in t_j},n_{ij}\rangle+d_{ij}\leq-\epsilon_{ij}/2.
\end{align*}
Clearly, $\mathcal{P}^{t_i\cup t_j}\leq\mathcal{L}_{ij}(p_{ij}/2,x_{i(k)\in t_i},x_{j(k)\in t_j})<\infty$. 
On the other hand, if $x\in\CObs$, then there exists a non-disjoint pair $t_i,t_j$, so for any separating plane $p_{ij}$ there exists $\langle x_{i(k)},n_{ij}\rangle+d_{ij}\leq0$ or $x_{j(k)},n_{ij}\rangle+d_{ij}\geq0$, leading to $\mathcal{P}^{t_i\cup t_j}=\infty$.
Finally, at any feasible solution $p_{ij}$, we must have $\|n_{ij}\|>0$ because otherwise, we have:
\begin{align*}
\mathcal{L}_{ij}=\sum_{k=1}^3\PXInv{d_{ij}}+\sum_{k=1}^3\PXInv{-d_{ij}}=\infty.
\end{align*}
We have thus established~\prettyref{prop:A} for each $\mathcal{P}^{t_i\cup t_j}$ and thus $\mathcal{P}$.

\TE{\prettyref{prop:B}} This follows from the inverse function theorem~\citep{dontchev2009implicit}, the smoothness of problem data $\mathcal{L}_{ij}$, and strictly convexity of $\mathcal{L}_{ij}$.

\TE{\prettyref{prop:C}} We can simply define $\mathcal{A}(x)=\{\langle t_i,t_j\rangle|t_i\neq t_j\}$ such that every pair of disjoint triangles $\langle t_i,t_j\rangle$ appears in exactly one term of $\mathcal{P}^{t_i\cup t_j}$. Thus, we only need to verify that $f_{i(k)\in t_i}^{t_i\cup t_j}\in\mathcal{F}_{t_j\to t_i}$ and the case with $f_{j(k)\in t_j}^{t_i\cup t_j}$ is symmetric. By the implicit function theorem, we can derive the analytic formula:
\begin{align*}
f_{i(k)\in t_i}^{t_i\cup t_j}=\frac{n_{ij}}{(\langle x_{i(k)},n_{ij}\rangle+d_{ij})^2}.
\end{align*}
Suppose $n_{ij}=\alpha(a-b)$ for some $a\in\CH(x_{i(k)\in t_i})$, $b\in\CH(x_{j(k)\in t_j})$, and $\alpha>0$, then we have $f_{i(k)\in t_i}^{t_i\cup t_j}=\alpha(a-b)/(\langle x_{i(k)},n_{ij}\rangle+d_{ij})^2\in\mathcal{F}_{t_j\to t_i}$. Therefore, we can in turn prove the sufficient condition that $n_{ij}=\alpha(a-b)$. Due to the optimality of $\mathcal{L}_{ij}$ with respect to $p_{ij}$, we have:
\begin{align*}
0=&\FPP{\mathcal{L}_{ij}}{n_{ij}}=
12\frac{n_{ij}}{\|n_{ij}\|(1-\|n_{ij}\|)^2}-
\sum_{k=1}^3\frac{x_{i(k)}}{(\langle x_{i(k)},n_{ij}\rangle+d_{ij})^2}+
\sum_{k=1}^3\frac{x_{j(k)}}{(\langle x_{j(k)},-n_{ij}\rangle-d_{ij})^2}\\
0=&\FPP{\mathcal{L}_{ij}}{d_{ij}}=-\sum_{k=1}^3\frac{1}{(\langle x_{i(k)},n_{ij}\rangle+d_{ij})^2}+
\sum_{k=1}^3\frac{1}{(\langle x_{j(k)},-n_{ij}\rangle-d_{ij})^2}.
\end{align*}
From the above two equations, we can conclude that $n_{ij}=\alpha(a-b)$ by defining:
\begin{align*}
\alpha=&\frac{\|n_{ij}\|(1-\|n_{ij}\|)}{12}\sum_{k=1}^3\frac{1}{(\langle x_{i(k)},n_{ij}\rangle+d_{ij})^2}>0\\
a=&\left[\sum_{k=1}^3\frac{x_{i(k)}}{(\langle x_{i(k)},n_{ij}\rangle+d_{ij})^2}\right]/\left[\sum_{k=1}^3\frac{1}{(\langle x_{i(k)},n_{ij}\rangle+d_{ij})^2}\right]\in\CH(x_{i(k)\in t_i})\\
b=&\left[\sum_{k=1}^3\frac{x_{j(k)}}{(\langle x_{j(k)},-n_{ij}\rangle-d_{ij})^2}\right]/\left[\sum_{k=1}^3\frac{1}{(\langle x_{j(k)},-n_{ij}\rangle-d_{ij})^2}\right]\in\CH(x_{j(k)\in t_j}),
\end{align*}
thus all is proved.
\end{proof}
\begin{proof}[Proof of~\prettyref{lem:blending}]
\TE{\prettyref{prop:A}} Case I: When $\|x_\mathcal{I}-x_\mathcal{J}\|<d_1$, $\mathcal{P}_{d_1\to d_2}^{\mathcal{I}\cup\mathcal{J}}=\mathcal{P}_{d_1}^{\mathcal{I}\cup\mathcal{J}}$ which pertains~\prettyref{prop:A} since $\mathcal{P}_{d_1}^{\mathcal{I}\cup\mathcal{J}}$ pertains~\prettyref{prop:A} by assumption. Case II: When $\|x_\mathcal{I}-x_\mathcal{J}\|=d_1\geq R_\mathcal{I}+R_\mathcal{J}$, there are two sub-cases. Case II.a: If $x\in\CObs^{\mathcal{I}\cup\mathcal{J}}$, e.g. where two bounding spheres are just touching and the touching point lies on a common triangle, then $0\leq\mathcal{P}_{d_2}^{\mathcal{I}\cup\mathcal{J}}\leq\mathcal{P}_{d_1}^{\mathcal{I}\cup\mathcal{J}}=\infty$ where the first two inequalities are due to our assumption and the last equality is due to $\mathcal{P}_{d_1}^{\mathcal{I}\cup\mathcal{J}}$ pertaining~\prettyref{prop:A}, so $\mathcal{P}_{d_1\to d_2}^{\mathcal{I}\cup\mathcal{J}}=\mathcal{P}_{d_1}^{\mathcal{I}\cup\mathcal{J}}=\infty$. Case II.b: If $x\in\CFree^{\mathcal{I}\cup\mathcal{J}}$, then we have $0\leq\mathcal{P}_{d_2}^{\mathcal{I}\cup\mathcal{J}}\leq\mathcal{P}_{d_1}^{\mathcal{I}\cup\mathcal{J}}<\infty$ following the same reasoning as case II.a, so $0\leq\mathcal{P}_{d_1\to d_2}^{\mathcal{I}\cup\mathcal{J}}<\infty$. Case III: When $\|x_\mathcal{I}-x_\mathcal{J}\|>d_1\geq R_\mathcal{I}+R_\mathcal{J}$, then we must have $x\in\CFree^{\mathcal{I}\cup\mathcal{J}}$ and the same analysis as case II.b leads to $0\leq\mathcal{P}_{d_1\to d_2}^{\mathcal{I}\cup\mathcal{J}}<\infty$. Thus, we have verified that $\mathcal{P}_{d_1\to d_2}^{\mathcal{I}\cup\mathcal{J}}$ pertains~\prettyref{prop:A} in all cases.

\TE{\prettyref{prop:B}} This is due to~\prettyref{prop:B} in $\mathcal{P}_{d_1}^{\mathcal{I}\cup\mathcal{J}}$, $\mathcal{P}_{d_2}^{\mathcal{I}\cup\mathcal{J}}$, and the second differentiability of $\phi_{d_1\to d_2}$.

\TE{\prettyref{prop:C}} The force $f_{i\in\mathcal{I}}^{\mathcal{I}\cup\mathcal{J}}$ induced by $\mathcal{P}_{d_1\to d_2}^{\mathcal{I}\cup\mathcal{J}}$ takes the following form:
\begin{equation}
\begin{aligned}
\label{eq:force-blending}
f_{i\in\mathcal{I}}^{\mathcal{I}\cup\mathcal{J}}=&\underbrace{-(1-\phi_{d_1\to d_2}(x))\FPP{\mathcal{P}_{d_1}^{\mathcal{I}\cup\mathcal{J}}}{x_{i\in\mathcal{I}}}}_{\text{term I}}
\underbrace{-\phi_{d_1\to d_2}(x)\FPP{\mathcal{P}_{d_2}^{\mathcal{I}\cup\mathcal{J}}}{x_{i\in\mathcal{I}}}}_{\text{term II}}\\
&\underbrace{-(\mathcal{P}_{d_2}^{\mathcal{I}\cup\mathcal{J}}-\mathcal{P}_{d_1}^{\mathcal{I}\cup\mathcal{J}})\frac{\phi'((\|x_{\mathcal{I}}-x_{\mathcal{J}}\|-d_1)/(d_2-d_1))}{(d_2-d_1)\|x_{\mathcal{I}}-x_{\mathcal{J}}\||\mathcal{I}|}(x_\mathcal{I}-x_\mathcal{J})}_{\text{term III}}.
\end{aligned}
\end{equation}
There are three terms and we show that each term belongs to $\mathcal{F}_{\mathcal{J}\to\mathcal{I}}$. For term I, we know that $-\FPPR{\mathcal{P}_{d_1}^{\mathcal{I}\cup\mathcal{J}}}{x_{i\in\mathcal{I}}}\in\mathcal{F}_{\mathcal{J}\to\mathcal{I}}$ since $\mathcal{P}_{d_1}^{\mathcal{I}\cup\mathcal{J}}$ satisfies~\prettyref{prop:C}. Since the coefficient $(1-\phi_{d_1\to d_2}(x))\geq0$ and $\mathcal{F}_{\mathcal{J}\to\mathcal{I}}$ is a cone, we conclude that term I is zero or belongs to $\mathcal{F}_{\mathcal{J}\to\mathcal{I}}$. The same reasoning applies to term II. For term III, $x_\mathcal{I}-x_\mathcal{J}$ belongs to $\mathcal{F}_{\mathcal{J}\to\mathcal{I}}$. Since $\mathcal{P}_{d_2}^{\mathcal{I}\cup\mathcal{J}}\leq\mathcal{P}_{d_1}^{\mathcal{I}\cup\mathcal{J}}$ by our assumption, the remaining coefficient is non-negative, thus term III is zero or belongs to $\mathcal{F}_{\mathcal{J}\to\mathcal{I}}$. Finally, at least one of term I or term II is non-zero, so we conclude that $f_{i\in\mathcal{I}}^{\mathcal{I}\cup\mathcal{J}}\in\mathcal{F}_{\mathcal{J}\to\mathcal{I}}$ and all is proved.
\end{proof}
\begin{proof}[Proof of~\prettyref{cor:blending}]
We first show that~\prettyref{eq:centered-potential} is correct. The objective function in~\prettyref{eq:centered-potential} is derived by replacing all $x_{i(k)}$ and $x_{j(k)}$ in $\mathcal{L}_{ij}$ with the center points $x_{t_i}$ and $x_{t_j}$, respectively. By symmetry, the optimal separating plane must be the middle surface between $x_{t_i}$ and $x_{t_j}$, taking the following form:
\begin{align*}
n_{ij}=\alpha(x_{t_i}-x_{t_j})\text{ and }
d_{ij}=-\alpha/2\langle x_{t_i}-x_{t_j},x_{t_i}+x_{t_j}\rangle.
\end{align*}
Plugging the optimal separating plane and solving for $\alpha$ leads to~\prettyref{eq:centered-potential}. Next, we show that all three assumptions in~\prettyref{lem:blending} hold. In fact, there are only two non-trivial assumptions. We first show that $\mathcal{P}_c^{t_i\cup t_j}\leq\mathcal{P}^{t_i\cup t_j}$ when $\|x_\mathcal{I}-x_\mathcal{J}\|\geq d_1$. Let use denote $p_{ij}^\star$ as the optimal separating plane for $\mathcal{P}^{t_i\cup t_j}$, then we have the following inequality:
\begin{align*}
\mathcal{P}_c^{t_i\cup t_j}=&
\argmin{p_{ij}}\left[12\PXInv{1-\|n_{ij}\|}+\sum_{k=1}^3\PXInv{\langle x_{t_i},n_{ij}\rangle+d_{ij}}+
\sum_{k=1}^3\PXInv{\langle x_{t_j},-n_{ij}\rangle-d_{ij}}\right]\\
\leq&12\PXInv{1-\|n_{ij}^\star\|}+\sum_{k=1}^3\PXInv{\langle x_{t_i},n_{ij}^\star\rangle+d_{ij}^\star}+
\sum_{k=1}^3\PXInv{\langle x_{t_j},-n_{ij}^\star\rangle-d_{ij}^\star}\\
\leq&12\PXInv{1-\|n_{ij}^\star\|}+\sum_{k=1}^3\PXInv{\langle x_{i(k)},n_{ij}^\star\rangle+d_{ij}^\star}+
\sum_{k=1}^3\PXInv{\langle x_{j(k)},-n_{ij}^\star\rangle-d_{ij}^\star}\\
=&\mathcal{P}^{t_i\cup t_j},
\end{align*}
where the first inequality is due to optimality of $\mathcal{P}_c^{t_i\cup t_j}$, the second inequality is due to the convexity of function $\PXInvR{\langle\bullet,n_{ij}^\star\rangle+d_{ij}^\star}$ and $\PXInvR{\langle\bullet,-n_{ij}^\star\rangle-d_{ij}^\star}$. We then show that $\mathcal{P}_c^{t_i\cup t_j}$ satisfies~\prettyref{eq:non-prehensile}. The force on any $x_{i(k)}$ takes the following form:
\begin{align*}
f_{i\in\mathcal{I}}^{\mathcal{I}\cup\mathcal{J}}=\frac{4(x_{t_i}-x_{t_j})}{\|x_{t_i}-x_{t_j}\|^{5/2}}\left[1+\frac{1}{\|x_{t_i}-x_{t_j}\|^{1/2}}\right],
\end{align*}
which clearly belongs to $\mathcal{F}_{\mathcal{J}\to\mathcal{I}}$, thus all is proved.
\end{proof}
\begin{lemma}
\label{lem:BSH-well-behaved-A}
If $\epsilon>0$ then $\mathcal{P}_\text{BSH}^{\mathcal{I}\cup\mathcal{J}}$ pertains~\prettyref{prop:A} for any node pair $\mathcal{I},\mathcal{J}$ of two rigid bodies.
\end{lemma}
\begin{proof}
First, by induction from leaf to the root node, we can verify that $0\leq\mathcal{P}_\text{BSH}^{\mathcal{I}\cup\mathcal{J}}$. 
Second, suppose $x\in\CFree$, then all the pair-wise terms between leaf nodes $\mathcal{P}_\text{BSH}^{t_i\cup t_j}<\infty$. Further, for all the centered potential in~\prettyref{eq:centered-cluster-potential}, we have $\mathcal{P}_c^{\mathcal{I}\cup\mathcal{J}}<\infty$ because they are evaluated only when $R_\mathcal{I}+R_\mathcal{J}\leq\|x_\mathcal{I}-x_\mathcal{J}\|$. The root potential $\mathcal{P}_\text{BSH}^{\mathcal{I}\cup\mathcal{J}}$ is then derived by a finite number of blending and summation so we have $\mathcal{P}_\text{BSH}^{\mathcal{I}\cup\mathcal{J}}<\infty$. 
Third, at any $x\in\CObs$, there exists a non-disjoint pair $t_i,t_j$ belonging to the two rigid bodies. We will show the following two claims hold by induction from leaf to root:
\begin{itemize}
\item $\mathcal{P}_\text{BSH}^{\mathcal{I}\cup\mathcal{J}}=\infty$ at any node such that $t_i\subseteq\mathcal{I}$ and $t_j\subseteq\mathcal{J}$.
\end{itemize}
\TE{Base Step:} $\mathcal{P}_\text{BSH}^{t_i\cup t_j}=\mathcal{P}_{d_1\to d_2}^{t_i\cup t_j}=\infty$ by~\prettyref{cor:blending}. 

\TE{Inductive Step:} We assume our first claim holds for any $t_i\in\mathcal{I}_c\in C(\mathcal{I})$ and $t_j\in\mathcal{J}_c\in C(\mathcal{J})$. If $t_i\subseteq\mathcal{I}$ and $t_j\subseteq\mathcal{J}$, we must have $\|x_\mathcal{I}-x_\mathcal{J}\|\leq R_\mathcal{I}+R_\mathcal{J}$ due to the pair $t_i,t_j$ being non-disjoint. The children set satisfying $t_i\in\mathcal{I}_c\in C(\mathcal{I})$ and $t_j\in\mathcal{J}_c\in C(\mathcal{J})$ can always be found, so we have $\mathcal{P}_\text{BSH}^{\mathcal{I}\cup\mathcal{J}}=\mathcal{P}_{d_1}^{\mathcal{I}\cup\mathcal{J}}\geq\mathcal{P}_\text{BSH}^{\mathcal{I}_c\cup\mathcal{J}_c}=\infty$. 
\end{proof}

To prove that $\mathcal{P}_\text{BSH}^{\mathcal{I}\cup\mathcal{J}}$ pertains~\prettyref{prop:C}, i.e., the non-prehensile and non-vanishing property, we also need to use induction. To this end, we establish the non-prehensile property for an index subset as follows:
\begin{define}
\label{def:BSH-restricted}
The pairwise potential $\mathcal{P}_\text{BSH}^{\mathcal{I}\cup\mathcal{J}}$ pertains~\prettyref{prop:C} restricted to $\langle\mathcal{I},\mathcal{J}\rangle$ if, at every $x\in\CFree$, we can define a finite family of set pairs $\mathcal{A}_{\mathcal{I}\cup\mathcal{J}}(x)$ such that $\mathcal{P}_\text{BSH}^{\mathcal{I}\cup\mathcal{J}}=\sum_{\langle \mathcal{I}',\mathcal{J}'\rangle\in\mathcal{A}_{\mathcal{I}\cup\mathcal{J}}(x)}\mathcal{P}_\text{BSH}^{\mathcal{I}'\cup\mathcal{J}'}$, where every term $\mathcal{P}_\text{BSH}^{\mathcal{I}'\cup\mathcal{J}'}$ satisfy~\prettyref{eq:non-prehensile}. Further, for every pair of disjoint triangles $\langle t_i,t_j\rangle$ such that $t_i\in\mathcal{I}$ and $t_j\in\mathcal{J}$ or vice versa, we have $t_i\cup t_j\subseteq\mathcal{I}'\cup\mathcal{J}'$ for at least one $\langle \mathcal{I}',\mathcal{J}'\rangle\in\mathcal{A}_{\mathcal{I}\cup\mathcal{J}}(x)$.
\end{define}
\begin{lemma}
\label{lem:BSH-well-behaved-C}
If $\epsilon>0$ then $\mathcal{P}_\text{BSH}^{\mathcal{I}\cup\mathcal{J}}$ pertains~\prettyref{prop:C} restricted to $\langle\mathcal{I},\mathcal{J}\rangle$ for any node pair $\mathcal{I},\mathcal{J}$ of two rigid bodies.
\end{lemma}
\begin{proof}
At $x\in\CFree$, we show the following two claims by induction from leaf to root:
\begin{itemize}
\item The pairwise potential $\mathcal{P}_{d_1}^{\mathcal{I}\cup\mathcal{J}}\geq\mathcal{P}_\text{BSH}^{\mathcal{I}\cup\mathcal{J}}\geq\mathcal{P}_{d_2}^{\mathcal{I}\cup\mathcal{J}}$ at any node when $\|x_\mathcal{I}-x_\mathcal{J}\|\geq R_\mathcal{I}+R_\mathcal{J}$.
\item The pairwise potential $\mathcal{P}_\text{BSH}^{\mathcal{I}\cup\mathcal{J}}$ pertains~\prettyref{prop:C} restricted to $\langle\mathcal{I},\mathcal{J}\rangle$ for any node pair.
\end{itemize}
\TE{Base Step:} For the pairwise potential $\mathcal{P}_\text{BSH}^{t_i\cup t_j}=\mathcal{P}_{d_1\to d_2}^{t_i\cup t_j}$, we define $\mathcal{A}_{t_i\cup t_j}(x)=\{\langle t_i,t_j\rangle\}$, then~\prettyref{prop:C} and the fact that $\mathcal{P}_\text{BSH}^{t_i\cup t_j}\geq\mathcal{P}_c^{t_i\cup t_j}$ follows from~\prettyref{cor:blending}.

\TE{Inductive Step I:} We assume our first claim holds for all $\mathcal{I}_c\in C(\mathcal{I})$ and $\mathcal{J}_c\in C(\mathcal{J})$, i.e. $\mathcal{P}_\text{BSH}^{\mathcal{I}_c\cup\mathcal{J}_c}\geq\mathcal{P}_c^{\mathcal{I}_c\cup\mathcal{J}_c}$ when $\|x_{\mathcal{I}_c}-x_{\mathcal{J}_c}\|\geq R_{\mathcal{I}_c}+R_{\mathcal{J}_c}$. We first show that $\mathcal{P}_{d_2}^{\mathcal{I}\cup\mathcal{J}}\leq\mathcal{P}_{d_1}^{\mathcal{I}\cup\mathcal{J}}$ when $\|x_\mathcal{I}-x_\mathcal{J}\|\geq R_\mathcal{I}+R_\mathcal{J}$. We note that $\mathcal{I}\cup\mathcal{J}$ contains at least 3 triangles, otherwise we reduce to the base case, so there are at least 2 terms of form $\mathcal{P}_\text{BSH}^{\mathcal{I}_c\cup\mathcal{J}_c}$. Each such term has the following lower bound:
\begin{align*}
\mathcal{P}_\text{BSH}^{\mathcal{I}_c\cup\mathcal{J}_c}\geq
\mathcal{P}_c^{\mathcal{I}_c\cap\mathcal{J}_c}=12\left[1+\frac{1}{\sqrt{\|x_{\mathcal{I}_c}-x_{\mathcal{J}_c}\|}}\right]^2
\geq12\left[1+\frac{1}{\sqrt{\|x_\mathcal{I}-x_\mathcal{J}\|+R_\mathcal{I}+R_\mathcal{J}}}\right]^2.
\end{align*}
Here, the first inequality is due to our inductive condition and the fact that our BSH is a layered hierarchy by~\prettyref{def:BSH-restricted}, so that $\|x_\mathcal{I}-x_\mathcal{J}\|\geq R_\mathcal{I}+R_\mathcal{J}$ implies $\|x_{\mathcal{I}_c}-x_{\mathcal{J}_c}\|\geq R_{\mathcal{J}_c}+R_{\mathcal{J}_c}$. The second inequality is because $x_{\mathcal{I}_c}$ (resp. $x_{\mathcal{J}_c}$) is at most $R_\mathcal{I}$ (resp. $R_\mathcal{J}$) from $x_\mathcal{I}$ (resp. $x_\mathcal{J}$). Using the above lower bound, we derive the following estimate:
\begin{align*}
\mathcal{P}_{d_1}^{\mathcal{I}\cup\mathcal{J}}\geq&
24\left[1+\frac{1}{\sqrt{\|x_\mathcal{I}-x_\mathcal{J}\|+R_\mathcal{I}+R_\mathcal{J}}}\right]^2\\
\geq&24+\frac{48}{\sqrt{\|x_\mathcal{I}-x_\mathcal{J}\|+R_\mathcal{I}+R_\mathcal{J}}}+\frac{24}{\|x_\mathcal{I}-x_\mathcal{J}\|+R_\mathcal{I}+R_\mathcal{J}}\\
\geq&24+\frac{48}{\sqrt{2\|x_\mathcal{I}-x_\mathcal{J}\|}}+\frac{24}{2\|x_\mathcal{I}-x_\mathcal{J}\|}\\
\geq&12+\frac{24}{\sqrt{\|x_\mathcal{I}-x_\mathcal{J}\|}}+\frac{12}{\|x_\mathcal{I}-x_\mathcal{J}\|}
=12\left[1+\frac{1}{\sqrt{\|x_\mathcal{I}-x_\mathcal{J}\|}}\right]^2=\mathcal{P}_c^{\mathcal{I}\cup\mathcal{J}}=\mathcal{P}_{d_2}^{\mathcal{I}\cup\mathcal{J}},
\end{align*}
where we use the fact that $\|x_\mathcal{I}-x_\mathcal{J}\|\geq d_1=R_\mathcal{I}+R_\mathcal{J}$ in the third inequality. As a result, we have our first claim holds for $\mathcal{I}\cup\mathcal{J}$ by the definition of the blending~\prettyref{eq:blending}.

\TE{Inductive Step II:} We assume our second claim holds for all $\mathcal{I}_c\in C(\mathcal{I})$ and $\mathcal{J}_c\in C(\mathcal{J})$, i.e. $\mathcal{P}_\text{BSH}^{\mathcal{I}_c\cup\mathcal{J}_c}$ pertains~\prettyref{prop:C} restricted to $\langle\mathcal{I}_c,\mathcal{J}_c\rangle$. We show that $\mathcal{P}_\text{BSH}^{\mathcal{I}\cup\mathcal{J}}$ pertains~\prettyref{prop:C} restricted to $\langle\mathcal{I},\mathcal{J}\rangle$ by considering three cases. Case II.a: If $\|x_\mathcal{I}-x_\mathcal{J}\|\leq d_1$, then $\mathcal{P}_\text{BSH}^{\mathcal{I}\cup\mathcal{J}}=\mathcal{P}_{d_1}^{\mathcal{I}\cup\mathcal{J}}$ consists of terms of form $\mathcal{P}_\text{BSH}^{\mathcal{I}_c\cup\mathcal{J}_c}$ each satisfying~\prettyref{prop:C} by our inductive condition. Let us now define the finite family of set pairs by the union $\mathcal{A}_{\mathcal{I}\cup\mathcal{J}}(x)=\cup_{\mathcal{I}_c\in C(\mathcal{I}),\mathcal{J}_c\in C(\mathcal{J})}\mathcal{A}_{\mathcal{I}_c\cup\mathcal{J}_c}(x)$. It can be verified that this union is a disjoint union, and for every pair of disjoint triangles $\langle t_i\in\mathcal{I},t_j\in\mathcal{J}\rangle$, we have $t_i\cup t_j\subseteq\mathcal{I}'\cup\mathcal{J}'$ belongs to exactly one such $\mathcal{A}_{\mathcal{I}_c\cup\mathcal{J}_c}(x)$. Further, we have $\mathcal{P}_\text{BSH}^{\mathcal{I}\cup\mathcal{J}}=\sum_{\langle \mathcal{I}',\mathcal{J}'\rangle\in\mathcal{A}_{\mathcal{I}\cup\mathcal{J}}(x)}\mathcal{P}_\text{BSH}^{\mathcal{I}'\cup\mathcal{J}'}$, where each $\mathcal{P}_\text{BSH}^{\mathcal{I}'\cup\mathcal{J}'}$ satisfies~\prettyref{eq:non-prehensile} due to~\prettyref{prop:C} of the corresponding $\mathcal{P}_\text{BSH}^{\mathcal{I}_c\cup\mathcal{J}_c}$ by our inductive condition. We have thus verified~\prettyref{prop:C} of $\mathcal{P}_\text{BSH}^{\mathcal{I}\cup\mathcal{J}}$. Case II.b: If $\|x_\mathcal{I}-x_\mathcal{J}\|\geq d_2$, then $\mathcal{P}_\text{BSH}^{\mathcal{I}\cup\mathcal{J}}=\mathcal{P}_{d_2}^{\mathcal{I}\cup\mathcal{J}}=\mathcal{P}_c^{\mathcal{I}\cup\mathcal{J}}$ is a singled, centered potential. We trivially define $\mathcal{A}_{\mathcal{I}\cup\mathcal{J}}(x)=\{\langle\mathcal{I},\mathcal{J}\rangle\}$ then clearly every pair of disjoint triangles $t_i\cup t_j\subseteq\mathcal{I}\cup\mathcal{J}\in\mathcal{A}_{\mathcal{I}\cup\mathcal{J}}(x)$. Further, the induced force takes the following form:
\begin{align*}
f_{i\in\mathcal{I}}^{\mathcal{I}\cup\mathcal{J}}=\frac{12(x_\mathcal{I}-x_\mathcal{J})}{|\mathcal{I}|\|x_\mathcal{I}-x_\mathcal{J}\|^{5/2}}\left[1+\frac{1}{\|x_\mathcal{I}-x_\mathcal{J}\|^{1/2}}\right],
\end{align*}
which clearly belongs to $\mathcal{F}_{\mathcal{J}\to\mathcal{I}}$, thus we have verified all conditions in~\prettyref{prop:C} of $\mathcal{P}_\text{BSH}^{\mathcal{I}\cup\mathcal{J}}$. Case II.c: If $d_1<\|x_\mathcal{I}-x_\mathcal{J}\|<d_2$, then $\mathcal{P}_\text{BSH}^{\mathcal{I}\cup\mathcal{J}}$ is a blending of $\mathcal{P}_{d_1}^{\mathcal{I}\cup\mathcal{J}}$ in Case II.a and $\mathcal{P}_{d_2}^{\mathcal{I}\cup\mathcal{J}}$ in Case II.b, with strictly positive weights. We again trivially define $\mathcal{A}_{\mathcal{I}\cup\mathcal{J}}(x)=\{\langle\mathcal{I},\mathcal{J}\rangle\}$ to have every pair of disjoint triangles $t_i\cup t_j\subseteq\mathcal{I}\cup\mathcal{J}\in\mathcal{A}_{\mathcal{I}\cup\mathcal{J}}(x)$. The only condition we need to verify is that $\mathcal{P}_\text{BSH}^{\mathcal{I}\cup\mathcal{J}}$ satisfies~\prettyref{eq:non-prehensile} as a single term. We use a similar technique as in the proof of~\prettyref{lem:blending} by expanding the force term to get~\prettyref{eq:force-blending} where there are three terms. For term II, We have by the analysis in case II.b that $-\FPPR{\mathcal{P}_{d_2}^{\mathcal{I}\cup\mathcal{J}}}{x_{i\in\mathcal{I}}}\in\mathcal{F}_{\mathcal{J}\to\mathcal{I}}$ and $\phi_{d_1\to d_2}(x)>0$ is strictly positive, so term II belongs to $\mathcal{F}_{\mathcal{J}\to\mathcal{I}}$. Term III is zero or belongs to $\mathcal{F}_{\mathcal{J}\to\mathcal{I}}$ because $\mathcal{P}_{d_1}^{\mathcal{I}\cup\mathcal{J}}\geq\mathcal{P}_{d_2}^{\mathcal{I}\cup\mathcal{J}}$ by our first claim. For term I, we know from the analysis in Case II.a that:
\begin{align*}
-\FPP{\mathcal{P}_{d_1}^{\mathcal{I}\cup\mathcal{J}}}{x_{i\in\mathcal{I}}}=\sum_{\langle \mathcal{I}',\mathcal{J}'\rangle\in\mathcal{A}_{\mathcal{I}\cup\mathcal{J}}(x)}-\FPP{\mathcal{P}_\text{BSH}^{\mathcal{I}'\cup\mathcal{J}'}}{x_{i\in\mathcal{I}}},
\end{align*}
where each term $-\FPPR{\mathcal{P}_\text{BSH}^{\mathcal{I}'\cup\mathcal{J}'}}{x_{i\in\mathcal{I}}}$ is zero or belongs to $\mathcal{F}_{\mathcal{J}'\to\mathcal{I}'}\subseteq\mathcal{F}_{\mathcal{J}\to\mathcal{I}}$. Combined with the fact that the coefficient $(1-\phi_{d_1\to d_2}(x))>0$ is strictly positive, we conclude that term I is zero or belongs to $\mathcal{F}_{\mathcal{J}\to\mathcal{I}}$. As a result, we see that $\mathcal{P}_\text{BSH}^{\mathcal{I}\cup\mathcal{J}}$ satisfies~\prettyref{eq:non-prehensile} as a single term, so~\prettyref{prop:C} holds.
\end{proof}
\begin{proof}[Proof of~\prettyref{thm:BSH-well-behave}]
\prettyref{prop:A} and~\prettyref{prop:C} follows from~\prettyref{lem:BSH-well-behaved-A} and~\prettyref{lem:BSH-well-behaved-C}, respectively. \prettyref{prop:B} follows from the fact that $\mathcal{P}_\text{BSH}^{\mathcal{I}\cup\mathcal{J}}$ is derived by a finite number of blending between pairwise potentials, and all potentials and blending operators are twice differentiable.
\end{proof}
\begin{figure*}[ht]
\centering
\includegraphics[width=.9\linewidth]{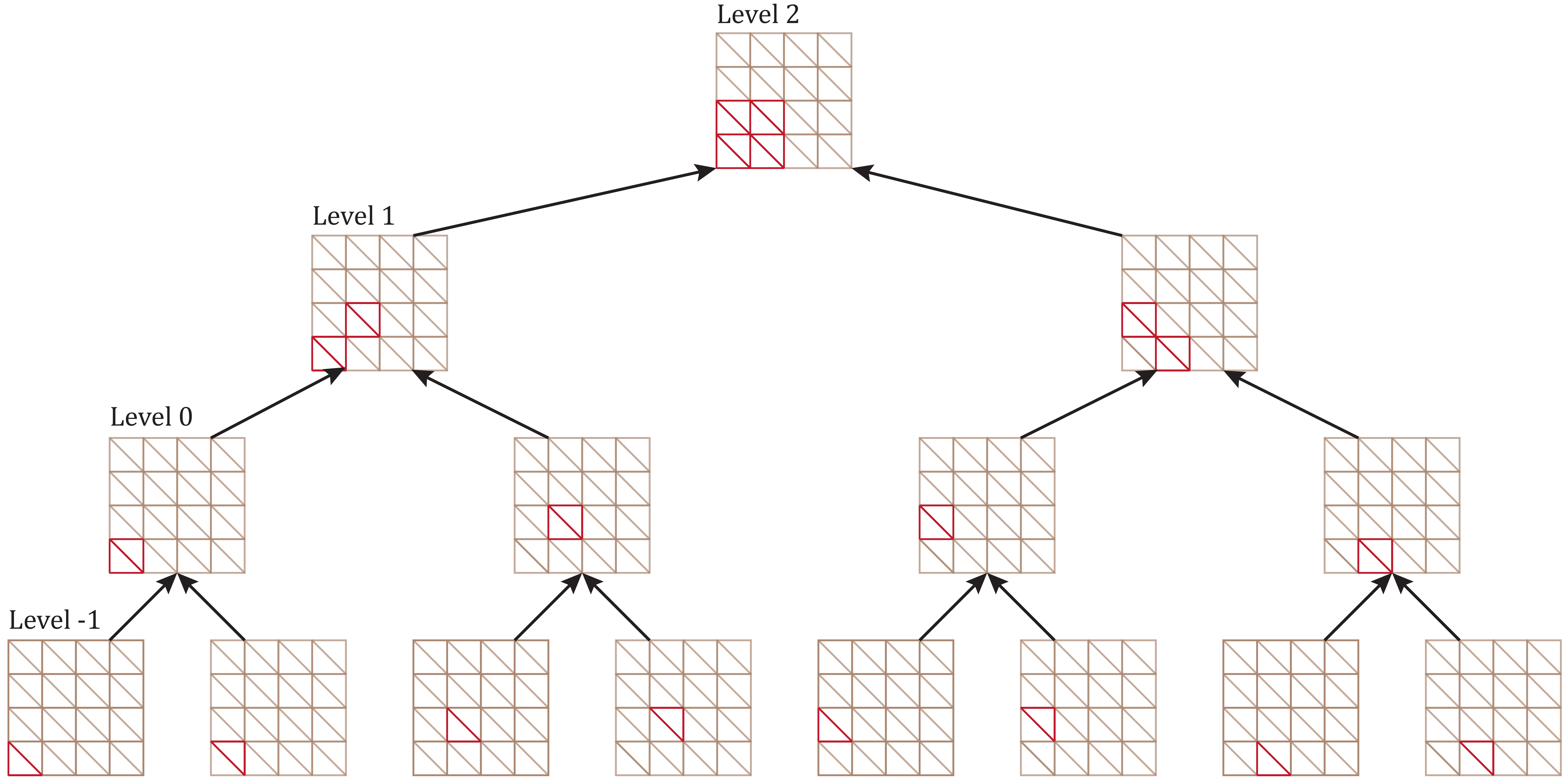}
\caption{\label{fig:BSH}A special BSH constructed for the uniform grid. The levels of the BSH are indexed from bottom up. The $-1$th level contains only leaves. Every $(2i+1)$th level ($i\geq0$) merges two diagonal, rectangular blocks. Therefore, every $(2i)$th level consists of a cubic mesh block of side length $2^i$.}
\end{figure*}
\subsection{\label{appen:complexity}Complexity Analysis for a Uniform Grid}
\begin{wrapfigure}{r}{0.5\textwidth}
\centering
\vspace{-10px}
\includegraphics[width=0.4\textwidth]{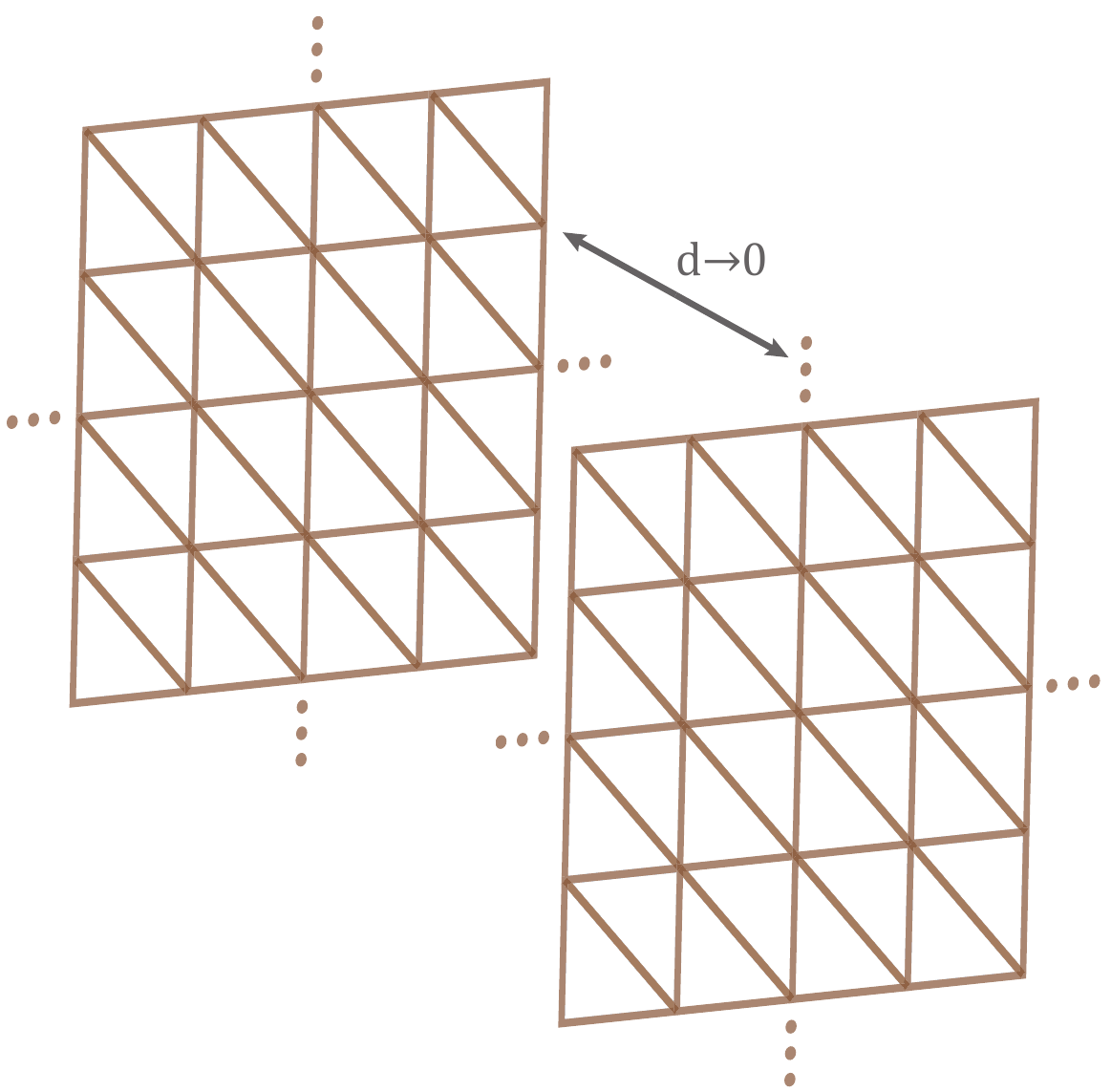}
\put(-100,150){$\mathcal{I}$}
\put(-30,105){$\mathcal{J}$}
\vspace{-5px}
\caption{\label{fig:uniform}\footnotesize{Two rigid bodies in the shape of a square uniform grid with infinitesimal distance $d\to 0$.}}
\vspace{-20px}
\end{wrapfigure}

We show that, in the special case of two rigid bodies in the shape of a square uniform grid with infinitesimal distace to each other, as illustrated in~\prettyref{fig:uniform}, the cost of evaluating $\mathcal{P}_\text{BSH}$ is $O(T)$.
Here we assume a 2D uniform grid with $N^2$ grid cells so that $T=O(N^2)$. Without a loss of generality, we assume the grid size is 1. Further, for ease of analysis, we adopt a special construction of BSH as illustrated in~\prettyref{fig:BSH}. It is easy to see that the bounding sphere radius of nodes in each level is all the same, which is denoted as $r_i$. We have the following results for estimating $r_i$, which can be derived directly from the construction of $\mathcal{P}_\text{BSH}^{\mathcal{I}\cup\mathcal{J}}$:
\begin{itemize}
\item $r_{-1}=\sqrt{5}/3\quad r_0=\frac{\sqrt{2}+2\sqrt{5}}{6}$
\item $r_{2i-1}=r_{2i}\leq 2^{i-1/2}C_r\quad\forall i>0$
with $C_r=\frac{1+\sqrt{10}}{3}$
\end{itemize}
To derive the second property, note that the tightest bounding sphere for the geometry of a $(2i)$-level node is $2^{i-1/2}$. This implies that the tightest bounding sphere for the geometry of a $0$-level node is $1/\sqrt{2}$. However, the actual $r_0=(\sqrt{1}+\sqrt{10})/3$, so the bounding sphere is unnecessarily scaled by $C_r$. By induction, we can prove that we can scale all tightest bounding spheres by $C_r$ accordingly to satisfy~\prettyref{def:BSH}.

To analyze the cost of evaluating $\mathcal{P}_\text{BSH}^{\mathcal{I}\cup\mathcal{J}}$, we note that, the cost reduces to evaluating a series of interaction terms either using centered potential ($\mathcal{P}_c^{\mathcal{I}\cup\mathcal{J}}$) or between leaf nodes ($\mathcal{P}^{t_i\cup t_j}$). Further by the recursive definition and the special structure of our BSH, the interaction terms are always computed between two nodes at the same level. Therefore, we can upper bound the number of interaction terms level by level.

\paragraph{Case I: $(2i)$-level Node}
We focus on the nodes at $(2i)$th level, which bounds a cubic mesh block of side length $2^i$. We can denote these mesh blocks as $\mathbb{B}^{2i}_{mn}$ indexed using subscript $mn$. In other words, $\mathbb{B}^{2i}_{mn}$ consists of all triangles with coordinates within $[m2^i,(m+1)2^i]\times[n2^i,(n+1)2^i]$. For two such blocks indexed by $\mathbb{B}^{2i}_{mn}\subset\mathcal{I}$ and $\mathbb{B}^{2i}_{m'n'}\subset\mathcal{I}$ on the two rigid bodies, we can define their distance as:
\begin{align*}
d(\mathbb{B}^{2i}_{mn},\mathbb{B}^{2i}_{m'n'})
=\max(|m-m'|,|n-n'|).
\end{align*}
Now suppose $d(\mathbb{B}^{2i}_{mn},\mathbb{B}^{2i}_{m'n'})\geq2$, we can be sure that $\mathbb{B}^{2i}_{mn}$ and $\mathbb{B}^{2i}_{m'n'}$ belong to different $(2i+2)$-level blocks. Specifically, we define $\mathbb{B}^{2i}_{mn}\subseteq\mathbb{B}^{2i+2}_{\bar{m}\bar{n}}$ and $\mathbb{B}^{2i}_{m'n'}\subseteq\mathbb{B}^{2i+2}_{\bar{m}'\bar{n}'}$ and we have the following relationship:
\begin{align}
\label{eq:condition-A}
d(\mathbb{B}^{2i+2}_{\bar{m}\bar{n}},\mathbb{B}^{2i+2}_{\bar{m}'\bar{n}'})
\geq\left\lfloor\frac{d(\mathbb{B}^{2i}_{mn},\mathbb{B}^{2i}_{m'n'})}{2}\right\rfloor.
\end{align}
Now let us define $\mathcal{I}=\mathbb{B}^{2i+2}_{\bar{m}\bar{n}}$ and $\mathcal{J}=\mathbb{B}^{2i+2}_{\bar{m}'\bar{n}'}$, we know that if the distance between the two cubic mesh blocks is sufficiently faraway, the potential term $\mathcal{P}^{\mathcal{I}\cup\mathcal{J}}$ reduces to the centered potential $\mathcal{P}_c^{\mathcal{I}\cup\mathcal{J}}$ can be computed via~\prettyref{eq:centered-cluster-potential} without utilizing any information from lower levels. A sufficient condition for this to happen is as follows:
\begin{align}
\label{eq:condition-B}
d(\mathbb{B}^{2i+2}_{\bar{m}\bar{n}},\mathbb{B}^{2i+2}_{\bar{m}'\bar{n}'})2^i\geq2r_{2i+2}(1+\epsilon).
\end{align}
Combining~\prettyref{eq:condition-A} and~\prettyref{eq:condition-B}, we know that if $d(\mathbb{B}^{2i}_{mn},\mathbb{B}^{2i}_{m'n'})\geq2\lceil2\sqrt{2}C_r(1+\epsilon)\rceil$, then the interaction between the two $(2i)$-level blocks would be handled by the two $(2i+2)$-level super-blocks. Therefore, in order to compute the contact potential, we only need to evaluate the interaction between $B^{2i}_{mn}$ and at most $(1+4\lceil2\sqrt{2}C_r(1+\epsilon)\rceil)^2$ blocks around it. The number of $(2i)$-level interaction terms of form $\mathcal{P}_c^{\mathcal{I}\cup\mathcal{J}}$ is:
\begin{align*}
O\left(\left\lceil\frac{N}{2^i}\right\rceil^2(1+4\lceil2\sqrt{2}C_r(1+\epsilon)\rceil)^2\right),
\end{align*}
where the first part $\lceil{N}/{2^i}\rceil^2$ is the number of $(2i)$-level blocks and the second part is the number of other $(2i)$-level blocks, with which an interaction term $\mathcal{P}_c^{\mathcal{I}\cup\mathcal{J}}$ needs to be calculated. 

\paragraph{Case II: $(2i-1)$-level Node}
We have finished analyzing the cost of $(2i)$-level node interactions. The case with $(2i-1)$-level node iterations ($i\geq1)$ is almost identical. Each $(2i-1)$-level node consists of half the triangles of a $(2i)$-level node, so we can assume the $(2i)$-level node $B^{2i}_{mn}$ has two children denoted as $\mathbb{B}^{2i-1,l}_{mn}$ and $\mathbb{B}^{2i-1,r}_{mn}$. Without a loss of generality, we only consider $\mathbb{B}^{2i-1,l}_{mn}$, which has the same bounding sphere center as that of $\mathbb{B}^{2i}_{mn}$. By the analysis of Case I, we know that, for all the children of $\mathbb{B}^{2i}_{m'n'}$ with $d(\mathbb{B}^{2i}_{mn},\mathbb{B}^{2i}_{m'n'})\geq2\lceil2\sqrt{2}C_r(1+\epsilon)\rceil$, their interaction with $\mathbb{B}^{2i-1,l}_{mn}$ would be taken care of at level $(2i+2)$. Again, we conclude that we only need to evaluate the interaction between $\mathbb{B}^{2i-1,l}_{mn}$ and the children of at most $(1+4\lceil2\sqrt{2}C_r(1+\epsilon)\rceil)^2$ $(2i)$-level nodes around it. The number of $(2i-1)$-level interaction terms of form $\mathcal{P}_c^{\mathcal{I}\cup\mathcal{J}}$ is again:
\begin{align*}
O\left(\left\lceil\frac{N}{2^i}\right\rceil^2(1+4\lceil2\sqrt{2}C_r(1+\epsilon)\rceil)^2\right).
\end{align*}

\paragraph{Case III: $-1$-level Leaf Node}
The case with leaf nodes is exactly the same as that of $(2i-1)$-level nodes. Each $0$-level node has two children at $-1$-level denoted as $\mathbb{B}^{-1,l}_{mn}$ and $\mathbb{B}^{-1,r}_{mn}$. Focusing on $\mathbb{B}^{-1,l}_{mn}$ and by the analysis of Case II, we know that, for all the children of $\mathbb{B}^{0}_{m'n'}$ with $d(\mathbb{B}^{0}_{mn},\mathbb{B}^{0}_{m'n'})\geq2\lceil2\sqrt{2}C_r(1+\epsilon)\rceil$, their interaction with $\mathbb{B}^{-1,l}_{mn}$ would be taken care of at level $2$. Therefore, we conclude that we only need to evaluate the interaction between $\mathbb{B}^{-1,l}_{mn}$ and the children of at most $(1+4\lceil2\sqrt{2}C_r(1+\epsilon)\rceil)^2$ $0$-level nodes around it. The number of $(2i-1)$-level interaction terms of form $\mathcal{P}^{t_i\cup t_j}$ is again:
\begin{align*}
O\left(N^2(1+4\lceil2\sqrt{2}C_r(1+\epsilon)\rceil)^2\right).
\end{align*}
Put everything together, the cost of evaluating $\mathcal{P}_\text{BSH}$ is:
\begin{align*}
\sum_{i=0}^{\lceil\log_2N\rceil}O\left(\left\lceil\frac{N}{2^i}\right\rceil^2\right)=O(N^2)=O(T).
\end{align*}
\subsection{\label{appen:fri}Frictional Contact Modeling}
A feature-complete potential function should be able to handle frictional contacts. To this end, we adopt the technique proposed by~\cite{li2020incremental,ye2024sdrs} and consider the contact potential between the pair of triangles $\mathcal{P}^{t_i\cup t_j}(x_{i(1)}^t,x_{i(2)}^t,x_{i(3)}^t,x_{j(1)}^t,x_{j(2)}^t,x_{j(3)}^t)$, where we write explicitly the six vertices related to this potential function. The negative gradient norm of this potential is the normal force applied on the corresponding vertices.~\cite{li2020incremental} proposes to model the contact potential as a tangential velocity damping term weight by the normal force magnitude. Put together, the frictional damping term takes the following form:
\begin{align*}
\mathcal{D}(x^{t+1},x^t,\delta t)
=\sum_{t_i\neq t_j}\left[\sum_{k=1}^3\lambda\left\|\frac{\mathcal{P}^{t_i\cup t_j}}{x_{i(k)}^t}\right\|D_\parallel(x_{i(k)}^{t+1},x_{i(k)}^t,\delta t)+
\sum_{k=1}^3\lambda\left\|\frac{\mathcal{P}^{t_i\cup t_j}}{x_{j(k)}^t}\right\|D_\parallel(x_{j(k)}^{t+1},x_{j(k)}^t,\delta t)\right],
\end{align*}
where the summation is taken over triangle pairs on different rigid bodies, $D_\parallel$ is the tangential velocity damping term, which penalizes the relative velocity between $t_i$ and $t_j$, and $\lambda$ is the frictional coefficient. However, the above formulation is not strictly second-order differentiable. To fix this problem,~\cite{ye2024sdrs} proposed a novel definition of $D_\parallel$, which does not penalize the relative velocity between $t_i$ and $t_j$. Instead, they penalize the relative velocity between the two triangles and the separating plane $p_{ij}$, assuming the plane is a physical object with zero-mass. We refer readers to their work for more details.

The original frictional damping term can only deals with frictions between triangles. However, our contact potential $\mathcal{P}_\text{BSH}^{\mathcal{I}\cup\mathcal{J}}$ is a hierarchical blending of potentials between triangles and centered potentials. To extend the frictional damping term to use our $\mathcal{P}_\text{BSH}^{\mathcal{I}\cup\mathcal{J}}$, we propose to disregard the centered potentials and only consider potentials between triangles. Specifically, we propose the following potential between a pair of triangles:
\begin{align}
\label{eq:leaf-potential-zero}
\begin{cases}
d_1=R_{t_i}+R_{t_j}\text{ and }d_2=(1+\epsilon)d_1\\
\mathcal{P}_{d_1}^{t_i\cup t_j}=\mathcal{P}^{t_i\cup t_j}\text{ and }
\mathcal{P}_{d_2}^{t_i\cup t_j}=0\\
\mathcal{P}_\text{local}^{t_i\cup t_j}=\mathcal{P}_{d_1\to d_2}^{t_i\cup t_j}
\end{cases}.
\end{align}
Compared with the potential between leaf nodes in~\prettyref{eq:leaf-potential},~\prettyref{eq:leaf-potential-zero} is not globally supported and vanishes when the distance between triangles is larger than $d_2$, denoted using subscript $\bullet_\text{local}$. This design choice would not cause gradient vanish because our normal contact potential $\mathcal{P}$ always provides non-vanishing gradient. In parallel, the benefit of using a locally supported function is that we can use a bounding volume hierarchy to quickly reject faraway triangles as done in~\cite{li2020incremental,ye2024sdrs}. \prettyref{eq:leaf-potential-zero} is plugged into the frictional damping term to yield our final formulation:
\begin{align*}
\mathcal{D}(x^{t+1},x^t,\delta t)
=\sum_{t_i\neq t_j}\left[\sum_{k=1}^3\lambda\left\|\frac{\mathcal{P}_\text{local}^{t_i\cup t_j}}{x_{i(k)}^t}\right\|D_\parallel(x_{i(k)}^{t+1},x_{i(k)}^t,\delta t)+
\sum_{k=1}^3\lambda\left\|\frac{\mathcal{P}_\text{local}^{t_i\cup t_j}}{x_{j(k)}^t}\right\|D_\parallel(x_{j(k)}^{t+1},x_{j(k)}^t,\delta t)\right].
\end{align*}
Intuitively, our frictional damping model assumes that two triangles can only impose frictional damping forces on each other if their distance is less than $d_2$. Otherwise, the two triangles can only impose normal forces, but not frictional damping forces. This design choice preserves the property of non-vanishing gradient, but also allows efficient evaluation. Finally, we emphasize that as $\mu\to0$, both our contact potential and frictional damping term converges to the exact frictional contact model with Coulomb friction.
\newpage
\subsection{\label{appen:IPC}Twice-Differentiability of IPC}
\begin{wrapfigure}{r}{0.5\textwidth}
\centering
\vspace{-10px}
\includegraphics[width=0.4\textwidth]{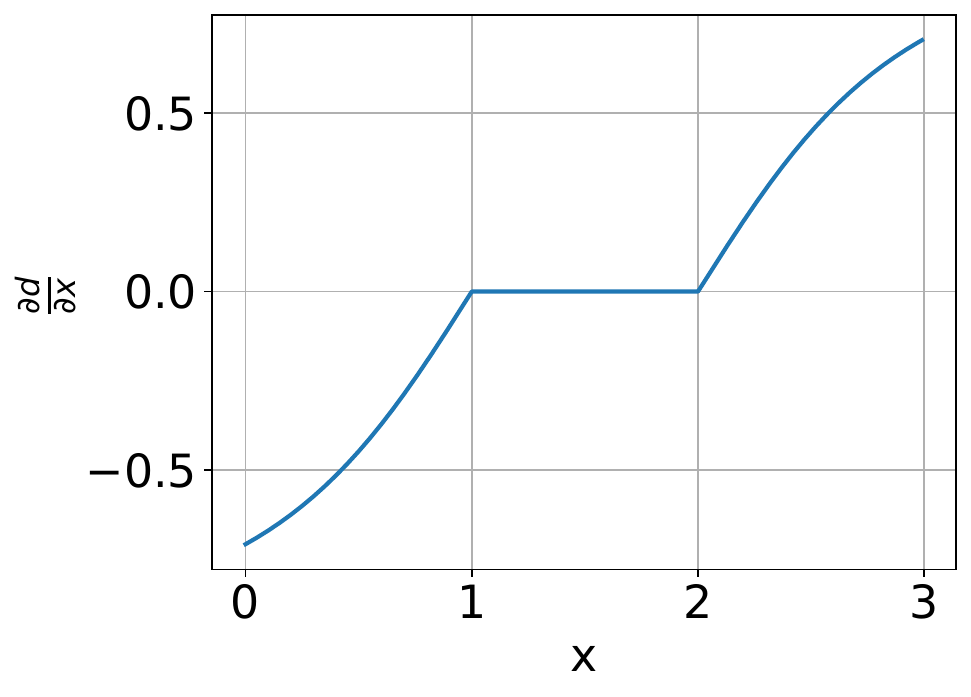}
\vspace{-5px}
\caption{\label{fig:gx-IPC}\footnotesize{We plot the value of $\partial d(x)/\partial x$ when $x\in[0,3]$ in a toy example under the IPC contact model.}}
\end{wrapfigure}
We show that the IPC contact model~\citep{li2020incremental} is differentiable but not twice differentiable. Let us consider a simple 2D case, where the only geometric primitive pairs that incur collision potential is between a point and a line-segment. Let us now assume the toy example with a single geometric primitive: a line segment with two end points located at $\TWO{1}{0}$ and $\TWO{2}{0}$ and a point moving on the line $\TWO{x}{1}$ with $x\in[0,3]$. The IPC potential for this toy example is formulated as: $\mathcal{P}(x)=-\log(d(x))$, where $d(x)$ is the distance between the point and the line-segment, parameterized by a single scalar $x$. Clearly, the differentiability of $\mathcal{P}$ relies on the differentiability of $d(x)$. As analyzed in~\cite{li2020incremental}, $d(x)$ is differentiable, so $\partial d(x)/\partial x$ is well-defined. In~\prettyref{fig:gx-IPC}, we plot the value of $\partial d(x)/\partial x$ when $x\in[0,3]$. Clearly, $\partial d(x)/\partial x$ is not a differentiable function, so we conclude that $\mathcal{P}(x)$ cannot be twice-differentiable. The non-smoothness is due to switching between different Voronoi regions. When $x\leq1$ or $x\geq2$, the closest point on the line-segment is a vertex. Instead, when $x\in(1,2)$, the closest point lies interior to the line segment.
\subsection{\label{appen:algorithm} Hierarchical Potential Blending}

Here we summarize the algorithm of hierarchical potential blending using BSH.

\begin{algorithm}
\caption{Hierarchical Potential Blending}
\begin{algorithmic}
\Function{Process\_Pair}{$\mathcal{I},\mathcal{J}$}
    \State 
$\mathcal{P}_{d_1}^{\mathcal{I}\cup\mathcal{J}}\leftarrow0$ \State $\mathcal{P}_{d_2}^{\mathcal{I}\cup\mathcal{J}}\leftarrow\mathcal{P}_c^{\mathcal{I}\cup\mathcal{J}}\triangleright{~\prettyref{eq:centered-cluster-potential}}$

    \If{$\mathcal{I}={t_i}$ is leaf and $\mathcal{J}={t_j}$ is leaf}
    \State
$\mathcal{P}_{d_1}^{\mathcal{I}\cup\mathcal{J}}\leftarrow\mathcal{P}^{{t_i}\cup{t_j}}$ $  \triangleright{~\prettyref{sec:slow}}$
    \State
Plug $\mathcal{P}_{d_1}, \mathcal{P}_{d_2}$ in ~\prettyref{eq:centered-cluster-potential} for $\mathcal{P}_{d_1\to d_2}$
     \State $\mathcal{P}_\text{BSH}^{\mathcal{I}\cup\mathcal{J}} \leftarrow \mathcal{P}_{d_1\to d_2}^{\mathcal{I}\cup\mathcal{J}}$
    \State \Return
    \EndIf
    
    \For{each pair $<\mathcal{I}_c,\mathcal{J}_c>$ in $<\mathcal{I},\mathcal{J}>$'s children}

        \State \Call{Process\_Pair}{$\mathcal{I}_c,\mathcal{J}_c$}
        \State $\mathcal{P}_{d_1}^{\mathcal{I}\cup\mathcal{J}} += \mathcal{P}_\text{BSH}^{\mathcal{I}_c\cup\mathcal{J}_c}$

    \EndFor
    Plug $\mathcal{P}_{d_1}, \mathcal{P}_{d_2}$ in ~\prettyref{eq:blending} for $\mathcal{P}_{d_1\to d_2}$
     \State $\mathcal{P}_\text{BSH}^{\mathcal{I}\cup\mathcal{J}} \leftarrow \mathcal{P}_{d_1\to d_2}^{\mathcal{I}\cup\mathcal{J}}$
    \State \Return
\EndFunction
\end{algorithmic}
\end{algorithm}
\subsection{\label{appen:experiment}Experimental Details}
In this section, we provide experimental details and extended evaluations.
\begin{table}[ht]
\centering
\small
\setlength{\tabcolsep}{6pt}
\begin{tabular}{lcccccc}
\toprule
Parameter & Billiards & Push & Ant-Push & Sort & Gather  &Gather-Bunny \\ 
\midrule
trajectory horizon $H$ & 100 & 200 & 240 & 700 & 300 &550\\ 
receding horizon $h$ & / & 48 & / & 16 & 32  &32\\  
potential coefficient $\mu$ & $1e^{-7}$   & $1e^{-6}$ & $1e^{-8}$   & $3e^{-8}$    & $3e^{-8}$ &$5e^{-9}$\\ 
$\alpha$ learning rate  & $3e^{-2}$  & $1e^{-2}$ & $3e^{-2}$ & $1e^{-2}$   & $1e^{-2}$ &$1e^{-2}$\\   
Adam $(\beta_1,\beta_2)$   & (0.3,0.5)  & (0.3,0.5) & (0.3,0.5) & (0.3,0.5)   & (0.3,0.5) &(0.3,0.5)\\ 
number of iterations & 400 & 50 & 100 & 100 & 60 &60\\ 
degrees of freedom & 96 & 12 & 22 & 66 & 66 &66\\ 
timesteps $\Delta t$ & 0.04 & 0.04 & 0.02 & 0.04 & 0.04 &0.04\\
\bottomrule
\end{tabular}
\vspace{2px}
\caption{\label{table:parameters} \small{Parameter settings for different benchmarks. Here, the trajectory horizon $H$ represents the total number of frames in the entire trajectory. For benchmarks requiring receding horizon execution, the receding horizon $h$ represents the number of frames in the sub-trajectory. The potential coefficient $\mu$ represents the contact energy coefficient, where a smaller $\mu$ indicates a more physically accurate contact mechanism. $\alpha$, the learning rate, denotes the step size for optimization, and the number of iterations specifies the total optimization steps. For all the benchmarks, we set the hyper-parameter of the Adam optimizer~\citep{kingma2014adam}, $\beta_1$ and $\beta_2$, to small values, which helps escaping from local minima. $\Delta t$ represents the timestep for each frame. Our simulator allows stable, penetration-free simulation even under relatively large timesteps.}}
\vspace{-10px}
\end{table}

\subsubsection{Baselines}
We compare our method against three state-of-the-art baselines.
The first is the IPC contact model~\citep{li2020incremental}, employed in the differentiable simulator of~\citep{huang2024differentiable}. However, due to the model's lack of twice-differentiability, the implicit function theorem does not apply. As a result,~\cite{huang2024differentiable} resort to a few iterations of gradient descent to approximately solve~\prettyref{eq:opt}. While this yields usable gradient information, it is well known that the gradients can vanish when the interacting geometric primitives are not in close proximity.
The second baseline is SDRS contact model ~\citep{ye2024sdrs}, it's a differentiable model with twice-differentiability. But similar to IPC, the gradient also vanish when the interacting convex hulls are far apart.
The third baseline is the Gradient Bundle (GB) method~\citep{suh2022bundled}, which addresses gradient vanishing through sampling, evaluated in practice via Monte Carlo methods. However, when primitives are far apart, the likelihood of sampling a non-zero gradient decreases significantly. Consequently, gradients obtained from GB can be both noisy and prone to vanishing with high probability.

\subsubsection{Benchmark Details}
We implement a full-featured rigid body simulator based on our novel contact model. Each benchmark scenario includes controlled objects and target objects. The controlled objects are actuated using a built-in PD controller, and the objective across all benchmarks is to manipulate collisions and contacts to move the target objects to their designated spatial positions. In all experiments, we use the following loss to measure the progress of optimizers:
\begin{align*}
\text{ReLU}(\|x_\text{COM}-x_\text{COM}^\star\|^2-\epsilon_\text{target}^2),
\end{align*}
where $x_\text{COM}$ and $x_\text{COM}^\star$ are the position and desired position of the center of mass of some target object. $\epsilon_\text{target}$ is the error coefficient, indicating that tasks are considered successful for certain objects when they are within $\epsilon_\text{target}$ of the goal. The statistics of benchmarks are summarized in~\prettyref{table:parameters}.

\paragraph{Billiards} In this task, the indices of the two target balls and their target locations are randomly selected. The objective is to control a distinct red ball so that it strikes the target balls through contact, moving them to their respective target positions. Each ball has 6 degrees of freedom, resulting in a total system of 96 degrees of freedom. We only control the initial horizontal positions and velocities of the red ball, corresponding to 4 control dimensions.

\paragraph{Push} In this benchmark, the task goal is to control a rod to push a box to the target region. The system consists of 2 objects with a total of 12 degrees of freedom. At each timestep, a continuous 6-dimensional control signal is generated to control the rod. The control signal at each timestep is obtained by solving a receding-horizon optimization problem. 

\paragraph{Ant-Push} In this benchmark, our goal is to drive the ant robot to move and push the box to the target position. The ant consists of a base, four large legs, and four small legs. The base includes 3 translational degrees of freedom and 1 rotational degree of freedom, while the upper legs and lower legs are connected using ball joints and hinge joints, respectively. As a result, the ant has a total of 16 degrees of freedom in the kinematic state. Combined with the 6 degrees of freedom of the box, the system has a total of 22 degrees of freedom in the kinematic state, of which we can control 12 degrees of freedom in the ant's legs. We use 4 accumulated sine wave signals to parameterize our controller for each degree of freedom of the ant's legs as done in~\cite{hu2019chainqueen}. In this case, the decision variables are the amplitude, frequency, and phase of the sine waves. 

\paragraph{Sort} In this benchmark, two types of cubes are mixed together on the ground. Use a rod to push each type of cube to its target location without mixing with each other. We set up 10 target cubes and one rod, so the system kinematic state has a total of 66 degrees of freedom and we can control the 3 translational degrees of freedom of the rod via PD controller. 

\paragraph{Gather} In this task, 10 cubes are randomly put on the ground. Use a rod to collect all the objects into one area. Again the task has a total of 66 degrees of freedom and we control 3 translational and 1 horizontal rotational degrees of freedom of the rod. To further validate the efficiency of our method, we can handle objects with more complex geometries in this scenario. We replace the cubes with bunnies and successfully complete the gather task, we call this benchmark Gather-Bunny.

\subsubsection{Ablation Study of Contact Property}
In ~\prettyref{fig:BSH-idea}, we discuss the various properties that a contact model needs to possess. Here, we will analyze the impact of these different properties on the contact model's performance in simulation and policy learning. Collision models without ~\prettyref{prop:A} would allow intersection between objects. It is well known that the distance function between objects is non-smooth when the objects are in collision. Therefore, if we take away ~\prettyref{prop:A}, then ~\prettyref{prop:B} will be violated automatically. ~\prettyref{prop:C}a  requires that normal collision forces between objects are pushing them apart, instead of pulling them together. Almost all existing collision models satisfy this property. We conduct an ablation study on~\prettyref{prop:C}b. Since the only difference between our method and SDRS model ~\citep{ye2024sdrs} is the additional satisfaction of~\prettyref{prop:C}b, the results of this ablation study can be observed in the comparison between our method and SDRS across various benchmarks in~\prettyref{sec:evaluation}. This demonstrates the necessity of satisfying this property for policy convergence when objects are far apart. 

We note that it is possible to take away~\prettyref{prop:B} alone and conduct an ablation study. To this end, we modify our ~\prettyref{eq:potential-slow} to use~\cite{li2020incremental} instead of~\cite{ye2024sdrs}. In other words, we combine~\cite{li2020incremental} with a globally supported barrier function instead of the original locally supported version. By doing so, we still have~\prettyref{prop:C}b but fails~\prettyref{prop:B}. We tested the results on the Gather and Push benchmarks, showing that without~\prettyref{prop:B}, the policy convergence speed is significantly slower or may even fail to converge completely.

\begin{figure*}[ht]
\centering
\scalebox{.73}{
\setlength{\tabcolsep}{1px}
\begin{tabular}{cc}
\includegraphics[height=.48\linewidth]{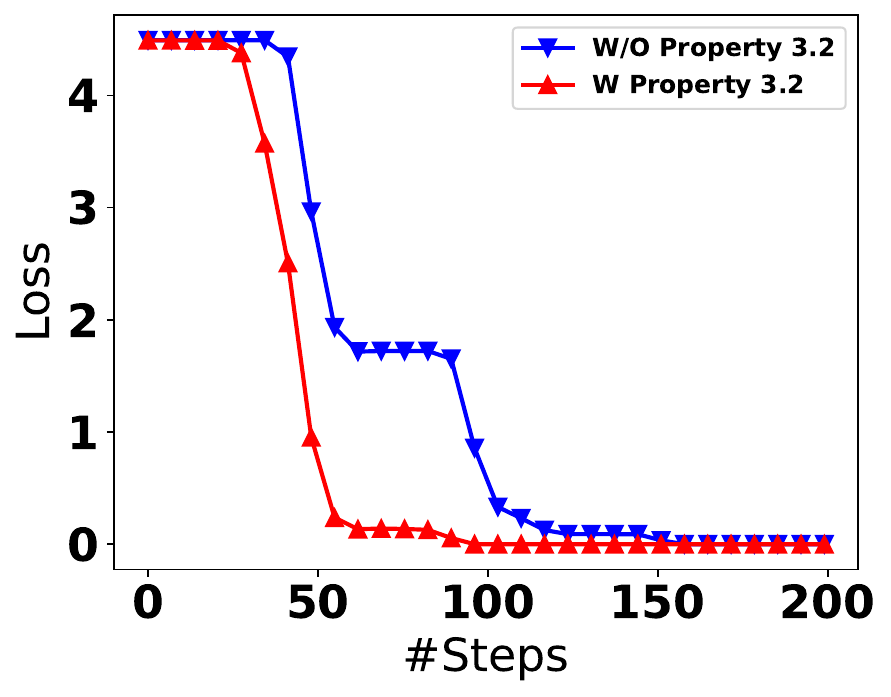}
\includegraphics[height=.48\linewidth]{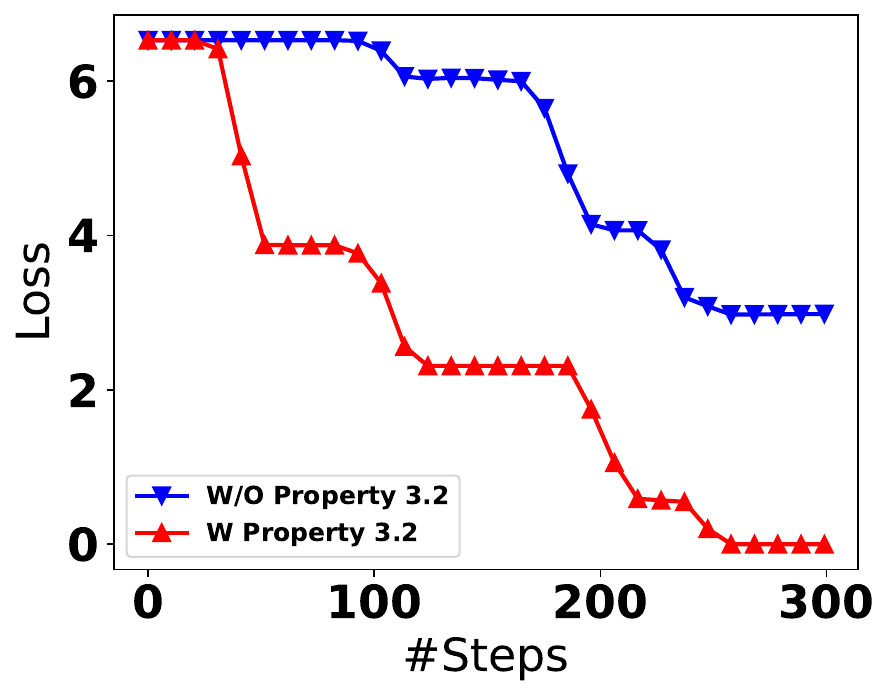}
\end{tabular}}
\caption{\label{fig:spider}\footnotesize{The convergency history with or without ~\prettyref{prop:B} on Push (left) and Gather (right) benchmark.}}
\end{figure*}

\newpage
\subsubsection{Computational Efficiency}
\begin{wrapfigure}{r}{0.5\textwidth}
\centering
\vspace{-10px}
\setlength{\tabcolsep}{5px}
\begin{tabular}{c}
\includegraphics[width=.95\linewidth,trim=2cm 4.5cm 0 8cm,clip]{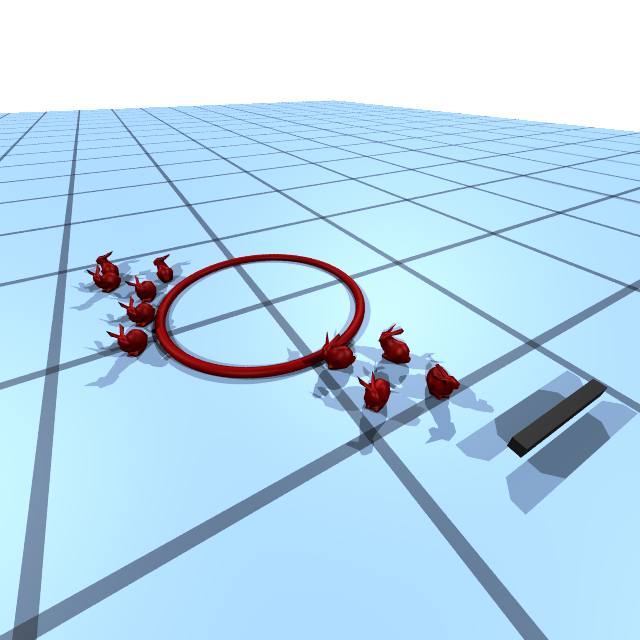}\\
\includegraphics[width=.95\linewidth,trim=2cm 4.5cm 0 8cm,clip]{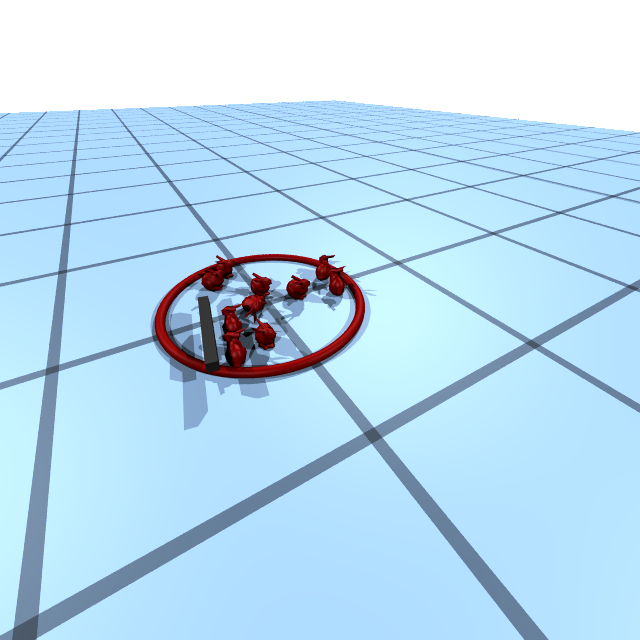}
\end{tabular}
\vspace{-5px}
\caption{\label{fig:bunny}\footnotesize{The start (top) and end (bottom) frame of the bunny-gather benchmark.}}
\vspace{-10px}
\end{wrapfigure}
As described in~\prettyref{sec:fast}, the brute-force evaluation of~\prettyref{eq:potential-slow} exhibits a time complexity of $O(T^2)$ with respect to the number of triangular facets, leading to rapidly increasing computational cost as the scene becomes more detailed. In contrast, our method achieves linear complexity in the special case outlined in~\prettyref{appen:complexity}, with the only overhead stemming from nested optimization and hierarchical blending, compared to the IPC model~\citep{li2020incremental}. To evaluate computational efficiency, we compared our method, brute-force computation, and IPC on the Billiards benchmark. We recursively subdivided the mesh of each ball to produce increasingly dense scenes, ranging from 512 to 6208 triangles in total. For each resolution, we performed full trajectory optimization to generate~\prettyref{fig:billiards}. With brute-force computation, the average time per frame increased significantly from 3.35s to 604.50s. In contrast, our BSH-assisted contact model saw a modest increase from 0.6s to 3.18s. These results empirically confirm the expected quadratic growth of brute-force computation and the near-linear scaling of our method. Interestingly, the IPC model~\citep{li2020incremental} maintained a consistent per-frame cost of approximately 0.5s across all resolutions. This is attributed to the optimizer requiring fewer Newton iterations as mesh resolution increases, offsetting the higher evaluation cost of the contact model. We have also compared our performance using more complex, non-convex geometric shapes. As illustrated in~\prettyref{fig:bunny}, we replace the cubes in our Gather benchmark with bunny meshes, with 2784 triangles in total. In this benchmark, named Gather-Bunny, we compare the average per-frame cost using brute-force computation, our method, and the IPC model, where the cost is 309.62s, 3.89s and 3.14s respectively.

\end{document}